\documentclass[a4paper]{article}

\usepackage{subfigure}
\usepackage{xspace}
\usepackage{graphicx}
\usepackage{amsmath}
\usepackage{amssymb}
\usepackage{amsmath}
\usepackage{nicefrac}

\usepackage[numbers]{natbib}
\usepackage{url}
\usepackage{color}
\usepackage{anysize}
\usepackage[color]{changebar}
\usepackage[affil-it]{authblk}
\usepackage{ntheorem}

\marginsize{1.5cm}{1.5cm}{1cm}{1cm}

\newtheorem{myconcept}{Concept}
\theoremstyle{break}
\newtheorem{mydef}{Definition}

\renewcommand{\cite}[1]{\citep{#1}}
\author[1]{Keyan Zahedi}
\author[1,2]{Nihat Ay}
\affil[1]{Information Theory of Cognitive Systems, Max Planck Institute for
  Mathematics in the Sciences, Leipzig, Saxony, Germany}
\affil[2]{%
Santa Fe Institute, 1399 Hyde Park Road, Santa Fe, New Mexico 87501, USA
}

\newcommand{\ASOCA}{\ensuremath{\mathrm{ASOC}_\mathrm{A}}\xspace}
\newcommand{\ASOCW}{\ensuremath{\mathrm{ASOC}_\mathrm{W}}\xspace}
\newcommand{\CA}{\ensuremath{\mathrm{C}_\mathrm{A}}\xspace}
\newcommand{\CW}{\ensuremath{\mathrm{C}_\mathrm{W}}\xspace}
\newcommand{\MCA}{\ensuremath{\mathrm{MC}_\mathrm{A}}\xspace}
\newcommand{\MCW}{\ensuremath{\mathrm{MC}_\mathrm{W}}\xspace}
\newcommand{\sgn}[1]{\ensuremath{{sign}\left(#1\right)}\xspace}
\title{Quantifying Morphological Computation}
\begin{document}
  \maketitle
  \section*{Abstract}
  The field of embodied intelligence emphasises the importance of the
  morphology and environment with respect to the behaviour of a cognitive
  system. The contribution of the morphology to the behaviour, commonly known as
  \emph{morphological computation}, is well-recognised in this community. We
  believe that the field would benefit from a formalisation of this concept as
  we would like to ask how much the morphology and the environment contribute to
  an embodied agent's behaviour, or how an embodied agent can maximise the
  exploitation of its morphology within its environment. In this work we derive
  two concepts of measuring morphological computation, and we discuss their
  relation to the Information Bottleneck Method. The first concepts asks how
  much the world contributes to the overall behaviour and the second concept
  asks how much the agent's action contributes to a behaviour. Various
  measures are derived from the concepts and validated in two experiments which
  highlight their strengths and weaknesses.
\bigskip\\
{\small \textbf{Keyworkds:} Information Bottleneck Method; Embodied Artificial Intelligence; Morphological Computation; Information Theory; Sensori-Motor Loop}
\bigskip\\
{\small The final version of this paper is published as open access article at:
  \emph{Zahedi K, Ay N. Quantifying Morphological Computation. Entropy. 2013;
    15(5):1887-1915, doi:10.3390/e15051887.}}

%%%%%%%%%%%%%%%%%%%%%%%%%%%%%%%%%%%%%%%%%%%%%%%%%%%%%%%%%%%%%%%%%%%%%%%%%%%%%%%%
%%%%%%%%%%%%%%%%%%%%%%%%%%%%%%%%%%%%%%%%%%%%%%%%%%%%%%%%%%%%%%%%%%%%%%%%%%%%%%%%
%%%   INTRODUCTION
%%%%%%%%%%%%%%%%%%%%%%%%%%%%%%%%%%%%%%%%%%%%%%%%%%%%%%%%%%%%%%%%%%%%%%%%%%%%%%%%
%%%%%%%%%%%%%%%%%%%%%%%%%%%%%%%%%%%%%%%%%%%%%%%%%%%%%%%%%%%%%%%%%%%%%%%%%%%%%%%%

\section{Introduction}

Morphological computation is discussed in various contexts, such as 
DNA computing and self-assembly \cite[see][for an
overview]{2007International-Conference-on-Morphological,Hauser2012Introduction-to-the-Special-Issue}.
This work is concerned with morphological computation in the field of embodied
intelligence. In this context it is often described as the trade-off between morphology
and control \cite{Pfeifer1999Understanding-intelligence}, which means that a
well-chosen morphology can reduce the amount of required control substantially.
Hereby, a morphology refers to the body of a system,
explicitly including all its physiological and physical properties (shape,
sensors, actuators, friction, mass distribution,
etc.)~\citep{Pfeifer2002Embodied-artificial-intelligence}. The consensus is
that
morphological computation is the contribution of the morphology and environment
to the behaviour, that cannot be assigned to a nervous system or a controller.
Theoretical work on describing morphological computation in this context has
been conducted by
\citep{Hauser2011Towards-a-theoretical-foundation,Fuchslin2012Morphological-Computation-and-Morphological}. 

The
following quote very nicely describes how the shape of an insect wing in flight
is not entirely determined by the muscular system, but by the interaction of the
wings' morphology with the environment:
\begin{quote}
  \emph{However, active muscular forces cannot entirely control the wing shape
    in flight. They can only interact dynamically with the aerodynamic and
    inertial forces that the wings experience and with the wing's own
    elasticity; the instantaneous results of these interactions are essentially
    determined by the architecture of the wing itself: its plan form and relief,
    the distribution and local mechanical properties of the veins, the local
    thickness and properties of the membrane, the position and form of lines of
    flexion. The interpretation of these characters is the core of functional
    wing morphology.}
  \citep[][see p.~188]{Wootton1992Functional-Morphology-of-Insect}
\end{quote}
The last sentence of this quote nicely summaries that the function of the wing
is determined by the interaction of the environment with the physical properties
of the wing. It is the quantification of this sort of contribution of the
morphology to the behaviour of a system that is in the focus of this work. The
difference to previous literature by
\citet{Paul2006Morphological-computation:-A-basis} and
\citet{Lundh2007A-quantification-of-the-morphological-computations}, who
investigated morphological computational with respect to either the actuation
(Paul) or the sensors (Lundh) only, is that we will measure morphological
computation of embodied agents acting in the sensori-motor loop, including both,
sensors and actuators.

To understand which aspects a measure must cover and which it should omit,
this paragraph will discuss two different artificial systems which show
morphological computation. To most vivid example in this context is
the Passive Dynamic Walker by \citet{McGeer1990Passive-dynamic-walking}. In this
example, a two-legged walking machine preforms a naturally appealing walking
behaviour, without any need of control, as a result of a well-chosen morphology
and environment. There is simply no computation available, and the walking
behaviour is the result of the gravity, the slope of the ground and the specifics
of the mechanical construction (weight \& length of the body parts, deviation of
the joints, etc.). If any parameter of the mechanics (morphology) or the slope
(environment) are changed, the walking behaviour will not persist. Hence, we will
only investigate morphological computation as an effect that emerges from the
interaction of the control system, the body and the environment, also known as
the sensori-motor loop (see next section). The Passive Dynamic Walker is, in
this context, understood as an embodied agent without actuation. The behaviour
of such a system is also discussed in the context of \emph{natural dynamics}~\citep{Carbajal2012Harnessing-Nonlinearities:-Behavior}.

One may argue that the Passive Dynamic Walker is a purely mechanical system, and
that speaking of morphological computation is an overstatement in this case.
This is a valid point of view, as purely mechanical system do not perform
calculations as we intuitively understand the term. Nevertheless, we claim that
it is not an overstatement in the case of the Passive Dynamic Walker, as this
system was explicitly built to simulate the morphological computation that is
present in the human walking behaviour. For this purpose, a mechanical system was
constructed that reflects the morphological properties of the lower half of a
human body as far as it is required to understand and model the principles
involved in human walking. Therefore, the Passive Dynamic Walker is allegoric
for morphological computation present in the locomotion of humans.

A second impressive example for morphological computation is BigDog
\citep{Marc-Raibert2008BigDog-the-Rough-Terrain-Quaduped}. This robot is a
four-legged walking machine, which is built as a companion for humans operating
in the field. The outstanding feature, with respect to this paper, is its
morphological design. Instead of classical electric motors, BigDog has hydraulic
actuators, which enable it to handle situations in which other walking machines
would fail and even suffer severe damage. Most impressive is the video, which
shows how BigDog handles a very slippery ground \citep{2012BigDog-on-YouTube}.
% \footnote{Watch
% \url{http://www.youtube.com/watch?v=W1czBcnX1Ww}, at about 1min 24s.}.
Watching the video, it seems impossible to program such a
balancing behaviour for an arbitrarily slippery ground. That BigDog is
nevertheless able to cope with it, is also the result of the hydraulic actuators
and the well-chosen morphological design in general, which are
both nicely documented in the video clip~\citep{2012BigDog-on-YouTube}.

The next step is to analyse, what both examples have in common, and in which
aspects the two applications differ. The first obvious difference is the amount
of available computation. Morphological computation is most often associated
with low computational power that leads to sophisticated behaviours
\citep[e.g.][]{Paul2006Morphological-computation:-A-basis}. This intuitive
understanding is well-reflected in the case of BigDog, as the on-board
computation accounts for only a fraction of BigDog's ability to cope with a
slippery ground. This intuitive understanding is not well-reflected in the case
of the Passive Dynamic Walker, as there is no computation available at all.
Therefore, measuring morphological computation in relation to the computational
complexity of the controller does not seem reasonable, as the amount of
computation used to generate the action does influence the amount of computation
conducted by the morphology.

There is something that both examples have in common, and which is well-suited
for a measure of morphological computation. In both cases, there is an observed
(coordinated) behaviour of the system that is not assigned to any control. In
the case of the Passive Dynamic Walker, this is obviously the case. In the case
of BigDog, it was stated above, that the behaviour of the robot on slippery
ground is not fully assigned to the control, but also assigned to the design of the
robot, and especially its hydraulic actuation. Therefore, a quantification
should capture how much of the behaviour is assigned to the morphology and
environment, and \emph{not} to the controller. Let us make
this point more clear. We are interested in the contribution of the morphology
and environment to the overall behaviour of a system. This is \emph{not}
well-measured in terms of controller complexity. This is also the reason why
we believe that BigDog is a good example for morphological computation,
although its control architecture is not known. 

This work is organised in the following way. The next section (see
Sec.~\ref{sec:sml}) describes the sensori-motor loop and its representation as a
causal graph, and thereby, presents the conceptual foundation that is required
for the remainder of this work. In the third section (see
Sec.~\ref{sec:measuring morphological computation}) we will then derive the
different measures for morphological computation. The section begins with a
summary of the intuitive understanding of morphological computation, which is
then followed by a discussion of two concepts to measure it. The section ends
with relating them to the Information Bottleneck Method by
\citet{Tishby1999The-information-bottleneck}. Two experiments to validate and
analyse the derived measures are conducted in the fourth section (see
Sec.~\ref{sec:experiments}). The fifth section (see Sec.~\ref{sec:discussion})
discusses the results and the final section (see Sec.~\ref{sec:conclusions})
concludes this work. All calculations are carried out in the appendix. The
results and derived measures are available as Mathematica 8 notebook file
\cite{Research2010Mathematica-Edition:-Version} in the electronic supplementary
information to this publication. 

%%%%%%%%%%%%%%%%%%%%%%%%%%%%%%%%%%%%%%%%%%%%%%%%%%%%%%%%%%%%%%%%%%%%%%%%%%%%%%%%
%%%%%%%%%%%%%%%%%%%%%%%%%%%%%%%%%%%%%%%%%%%%%%%%%%%%%%%%%%%%%%%%%%%%%%%%%%%%%%%%
%%%   SENSORI-MOTOR LOOP
%%%%%%%%%%%%%%%%%%%%%%%%%%%%%%%%%%%%%%%%%%%%%%%%%%%%%%%%%%%%%%%%%%%%%%%%%%%%%%%%
%%%%%%%%%%%%%%%%%%%%%%%%%%%%%%%%%%%%%%%%%%%%%%%%%%%%%%%%%%%%%%%%%%%%%%%%%%%%%%%%

\section{Sensori-Motor Loop}
\label{sec:sml}

\begin{figure}[t]
  \begin{center}
    \subfigure[Schematics of the sensori-motor loop]{
      \label{fig:sml concept}
        \includegraphics[height=4.25cm]{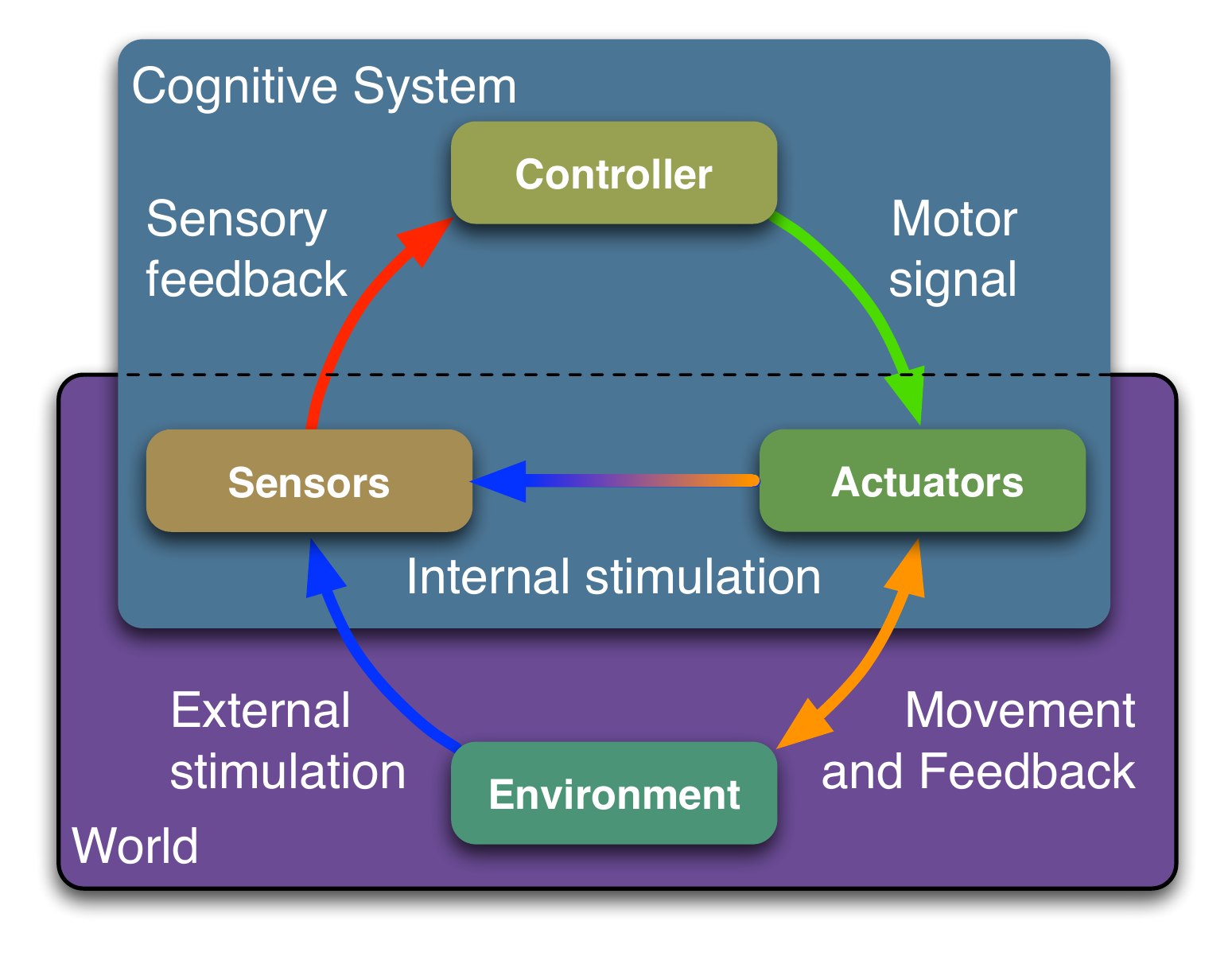}}
    \hspace*{2em}
    \subfigure[Representation of the sensori-motor loop as a causal graph]{
      \label{fig:sml full causal}
      \includegraphics[height=4.25cm]{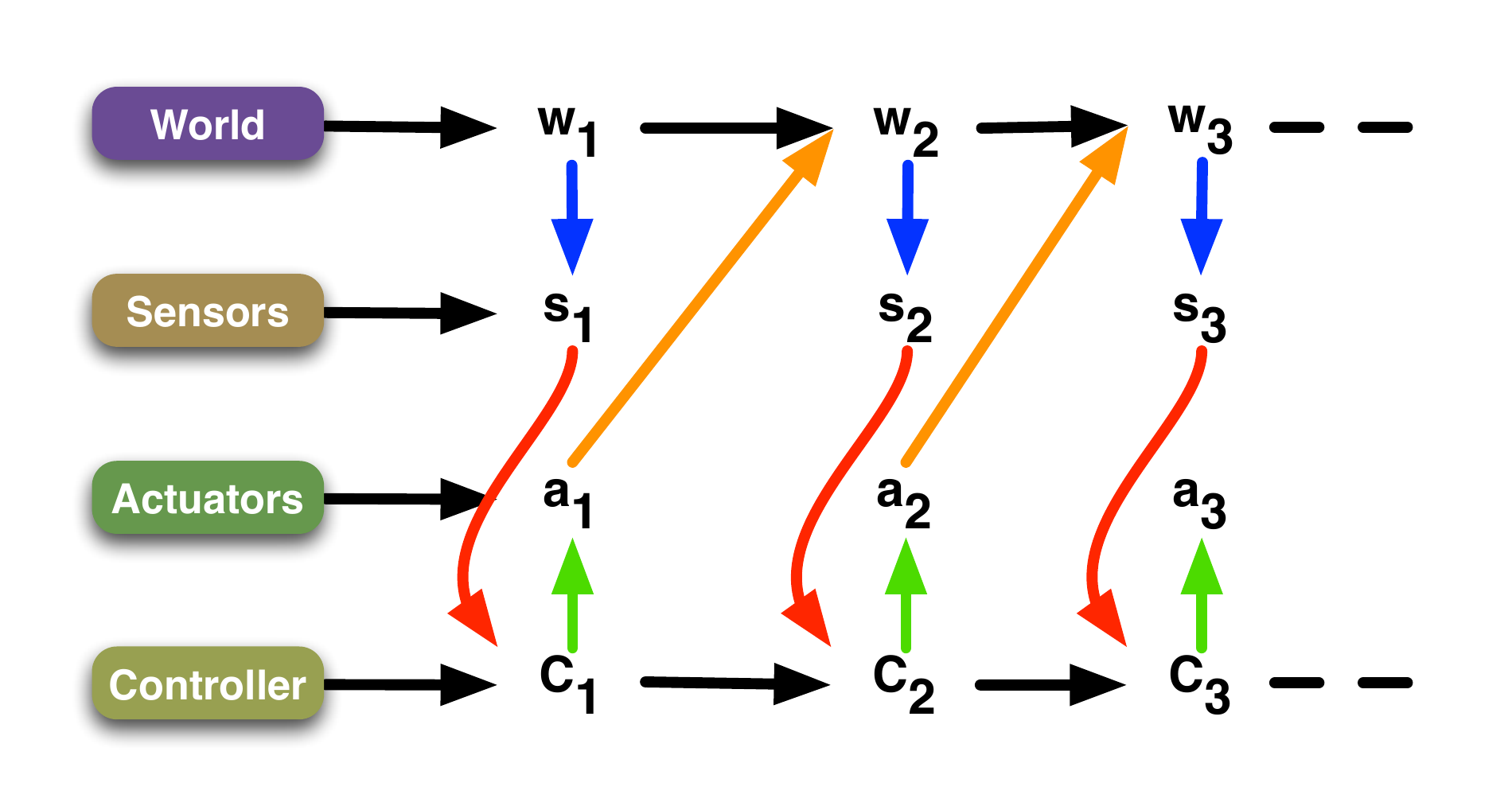}}
      \caption{Schematics and causal graph of the sensori-motor loop. The figure
        (a) shows the conceptual understanding of the sensori-motor loop. A
        cognitive system consists of a controller, a sensor and actuator
        system, and a body which is situated in an environment. The basic
        understanding is that the controller sends signals to the actuators
        which affect the environment. Information about the environment and also
        about internal states are sensed by the sensors, and the loop is closed
        when this information is passed to the controller. The figure (b) shows
        the representation of the sensori-motor loop as a causal graph. Here
        $w_t$ represents the world state at time $t$. The world is everything
        that is physical, i.e.~the environment and the morphology. The variables
        $s_t$ and $a_t$ are the signals provided by the sensors or passed to the
        actuators, respectively. They are not to be mistaken with the sensors
        and actuators, which are part of the morphology, and hence, part of the world.}
    \label{fig:sml}
  \end{center}
\end{figure}
To derive a quantification of morphological computation of an embodied system,
it requires a formal model of the sensori-motor loop. This will be briefly
summarised in the following paragraph (see also Fig.~\ref{fig:sml concept}). A
cognitive system consists of a brain or controller, which sends signals to the
system's actuators. The actuators affect the system's environment. We prefer the
notion of the system's \emph{Umwelt} \citep{Uexkuell1957A-Stroll-Through,Clark1996Being-There:-Putting,Zahedi2010Higher-coordination-with}, which
is the part of the system's environment that can be affected by the system, and
which itself affects the system. The state of the actuators and the
\emph{Umwelt} are not directly accessible to the cognitive system, but the loop
is closed as information about both, the \emph{Umwelt} and the actuators are
provided to the controller by the system's sensors. In addition to this general
concept, which is widely used in the embodied artificial intelligence community
\citep[see e.g.][]{Pfeifer2007Self-Organization-Embodiment-and} we introduce the
notion of \emph{world} to the sensori-motor loop, and by that we mean the
system's morphology and the system's \emph{Umwelt}. We can now distinguish
between the intrinsic and extrinsic perspective in this context. The world is everything that is
extrinsic from the perspective of the cognitive system, whereas the controller,
sensor and actuator signals are intrinsic to the system. 

The distinction between intrinsic and extrinsic is also captured in the
representation of the sensori-motor loop as a causal or Bayes'ian graph (see
Fig.~\ref{fig:sml full causal}). The random variables $C$, $A$, $W$, and $S$ refer to
the controller, actuator signals, world and sensor signals, and the directed
edges reflect causal dependencies between the random variables (see
\citep{Klyubin2004Organization-of-the-information-flow,Ay2008Information-Flows-in,Zahedi2010Higher-coordination-with}). Everything that
is extrinsic is captured in the variable $W$, whereas $S$, $C$, and $A$ are
intrinsic to the system. The random variables $S$ and $A$ are not to be mistaken
with the sensors and actuators. The variable $S$ is the output of the sensors,
which is available to the controller or brain, the action $A$ is the input that
the actuators take. Consider an artificial robotic system as an example. Then
the sensor state $S_t$ could be the pixel matrix delivered by some camera sensor
and the action $A_t$ could be a numerical value that is taken by some motor
controller to be converted in currents to drive a motor.

Throughout this work, we use capital letter ($X$, $Y$, \ldots) to denote random
variables, non-capital letter ($x$, $y$, \ldots) to denote a specific value that
a random variable can take, and calligraphic letters ($\mathcal{X}$,
$\mathcal{Y}$, \ldots) to denote the alphabet for the random variables. This
means that $x_t$ is the specific value that the random variable $X$ can take a
time $t\in\mathbb{N}$, and it is from the set $x_t\in\mathcal{X}$. Greek letters
refer to generative kernels, i.e.~kernels which describe an actual underlying
mechanism or a causal relation between two random variables. In the causal
graphs throughout this paper, these kernels are represented by direct
connections between the corresponding nodes. This notation is used to
distinguish generative kernels from others, such as the conditional probability
of $s_{t+1}$ given that $c_t$ was previously seen, denoted by $p(s_{t}|c_t)$,
which can be calculated or sampled, but which does not reflect a direct causal
relation between the two random variables $C_t$ and $S_{t+1}$ (see
Fig.~\ref{fig:sml full causal}).

We abbreviate the random variables for better comprehension in the remainder of
this work, as all measures consider random variables of consecutive time
indices. Therefore, we use the following notation. Random variables without any
time index refer to time index $t$ and hyphened variables to time index $t+1$.
The two variables $W,W'$ refer to $W_t$ and $W_{t+1}$.

We have now defined the conceptual framework in which we will derive the measures
for morphological computation. This is done in the next section, which starts
with an overview.

%%%%%%%%%%%%%%%%%%%%%%%%%%%%%%%%%%%%%%%%%%%%%%%%%%%%%%%%%%%%%%%%%%%%%%%%%%%%%%%%
%%%%%%%%%%%%%%%%%%%%%%%%%%%%%%%%%%%%%%%%%%%%%%%%%%%%%%%%%%%%%%%%%%%%%%%%%%%%%%%%
%%%   MORPHOLOGICAL COMPUTATION
%%%%%%%%%%%%%%%%%%%%%%%%%%%%%%%%%%%%%%%%%%%%%%%%%%%%%%%%%%%%%%%%%%%%%%%%%%%%%%%%
%%%%%%%%%%%%%%%%%%%%%%%%%%%%%%%%%%%%%%%%%%%%%%%%%%%%%%%%%%%%%%%%%%%%%%%%%%%%%%%%

\section{Measuring Morphological Computation}
\label{sec:measuring morphological computation}

\begin{figure}[t]
  \begin{center}
    \includegraphics[width=0.8\textwidth]{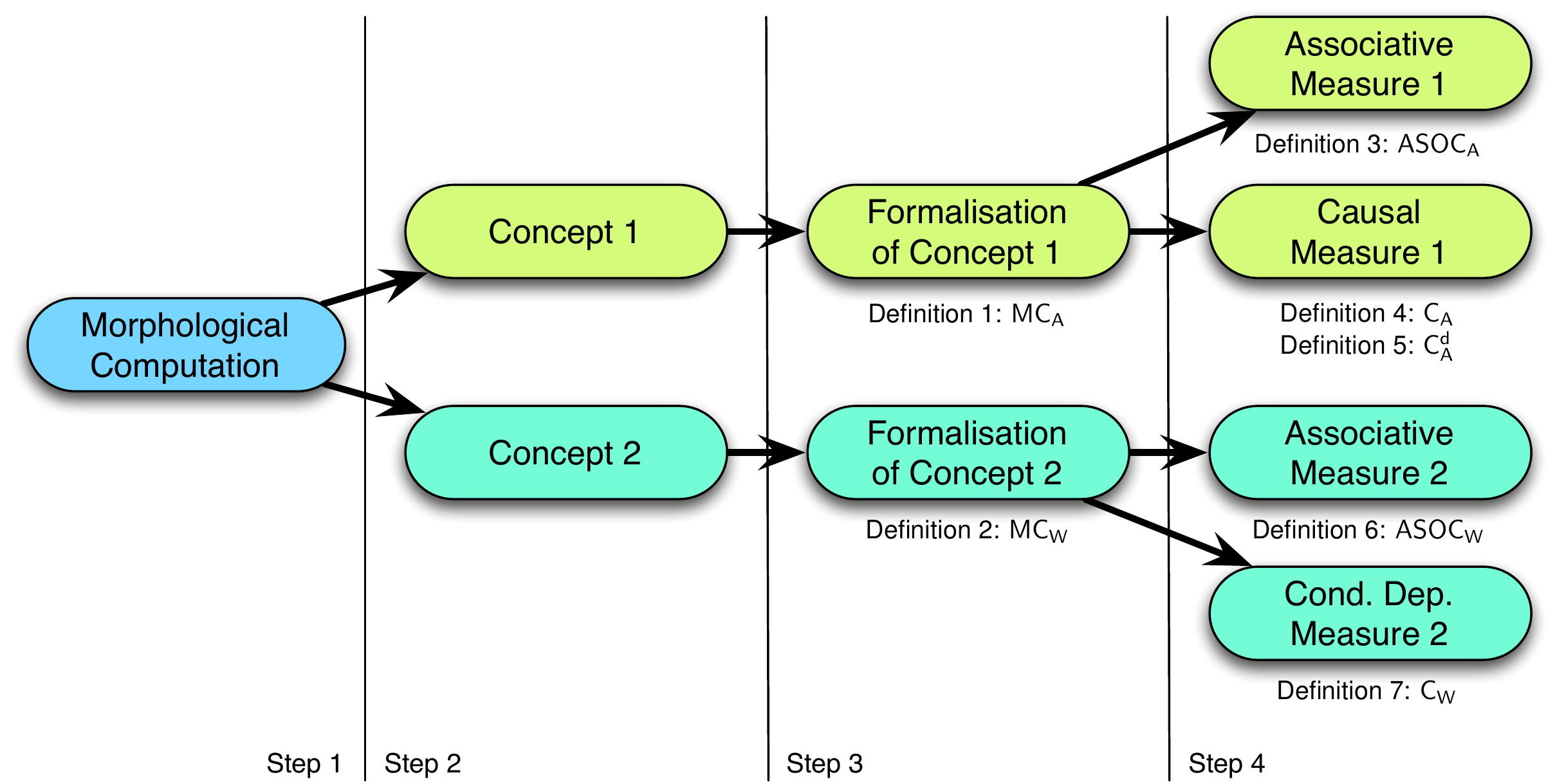}
  \end{center}
  \caption{Outline of the third section of this work, in which the
    measures for morphological computation are derived. The first step is
    to provide a description of morphological computation. From this description
    two concepts are presented in the second step,
    which are then formalised in the third step. The last step
    adapts the formalisations to measures which operate on intrinsically
    available information only.
  }
  \label{fig:outline}
\end{figure}

We will derive the measures for morphological computation in four steps (see
Fig.~\ref{fig:outline}):
\begin{enumerate}
  \item Descriptive definition of morphological computation.
  \item Framing of two concepts which follow from step 1.
  \item Formal definition of the two concepts given in step 2.
  \item Intrinsic adaptations of the definitions given in step 3.
\end{enumerate}

The fourth step is required because the two definitions given in the third step
require information about the world state $W$, which is generally not available.
This will be discussed in detail later in this section.

\subsection{Morphological Computation}

We understand morphological computation as the contribution of the morphology
and environment to the overall behaviour of an embodied system. This implies that
morphological computation must be studied in the sensori-motor loop, as it is
the result of the interaction of the controller, morphology and environment.

Measuring morphological computation means that an observer has
recorded the relevant data of an embodied agent's behaviour of interest and
post-hoc asks the question how much morphological computation was present in the
recoded sequence.

Measuring morphological computation does not mean to capture the complexity or
the quality of a behaviour. These are aspects of the behaviour of cognitive systems,
which are not handled by our measures.

\subsection{Concepts of measuring morphological computation}

In the following, we assume that an embodied agent is well-captured by the
sensori-motor loop as it was presented in the previous section (see
Fig.~\ref{fig:sml}). Explicitly, we
will use the term world in the way it was defined above, as the system's morphology and
\emph{Umwelt}.

Generally speaking, morphological computation is the effect of the current world
state $W$ on the next world state $W'$, which is not assignable to the action
$A$. Given the sensori-motor loop as we have
defined it above (see Fig.~\ref{fig:sml}), this cannot be measured in isolation,
as there is a path from $W$ to $W'$ which goes through the agent.
Hence, we need to measure the
effect of the action $A$ on the next world state $W'$ so that we are able to
deduce the effect of $W$ on $W'$ that does not include the pathway over $A$.
This has two implications. First, the resulting
value of such a measure is not an absolute value per se. In fact, it can only be
measured as the relation of the two effects $W\rightarrow W'$ and $A\rightarrow
W'$. Second, there are two ways of comparing the effects.

The first method of measuring morphological computation assumes that the
next world state $W'$ is only determined by the current world state $W$, which
is equivalent to assuming maximal morphological computation. Any measured effect
of the action $A$ on the next world state $W'$ displays itself in a reduction of
the resulting measurement. This is summarised in the following concept:

\begin{myconcept}[Negative effect of the action]
  Given a behaviour of interest of an embodied system, the amount of
  morphological computation is inversely proportional to the contribution of the
  actions of the system to the overall behaviour.
\end{myconcept}
This concept relates to control-dependent punishments 
discussed in literature to achieve high morphological computation
\citep[e.g.][]{Ruckert2012Stochastic-Optimal-Control}.

The second method of measuring morphological computation assumes that the
next world state $W'$ is only determined by the current action $A$, which
is equivalent to assuming no morphological computation. Any measured effect
of the current world state $W$ on the next world state $W'$ displays itself in
an increase of the resulting measurement. This is summarised in the following
concept:

\begin{myconcept}[Positive effect of the world]
  Given a behaviour of interest of an embodied system, the amount of
  morphological computation is proportional to the contribution of the world to
  the overall behaviour.
\end{myconcept}

\subsection{Formalising the concepts}

The next step is to formalise the two concepts presented above. The causal diagram
(see Fig.~\ref{fig:sml full causal}) shows that the world kernel $\alpha(w'|w,a) = p(w'|w,a)$
captures the influence of $W$ and $A$ on $W'$. Therefore, it is the basis for
our further considerations. If the action $A$ has no effect on the next world
state $W'$, then the world kernel reduces to $p(w'|w,a) = p(w'|w)$ and we would
state that the system shows maximal morphological computation. Analogously, if
the world state $W$ has no influence on the next world state $W'$, i.e.~$p(w'|w,a) =
p(w|a)$, we would state that the system shows no morphological computation.
Measuring the differences of the world kernel to the two conditional probability
distributions $p(w'|w)$ and $p(w'|a)$ leads us to the formalisations of the two
concepts discussed above. The Kullback-Leibler divergence measures the
differences of two probability distributions
\cite{Cover2006Elements-of-Information} and applying it to our scenario gives us
the definitions for the two concepts. Their formalisations are given below, such that the value
zero refers to no morphological computation and one refers to maximal morphological
computation.

\begin{mydef}[Morphological Computation as negative effect of the action]
  \label{def:MC_A}
  Let the random variables $A,W,W'$ denote the action, the current and the next
  world state of an embodied agent, which is described by the causal diagram
  shown in Figure~\ref{fig:sml full causal}. The quantification of the
  morphological computation as the negative effect of the action on the
  behaviour is then defined as
  \begin{align*}
    \MCA & := 1 - \frac{1}{\ln\vert\mathcal{W}\vert} D(p(w'|w,a)||p(w'|w))\\
           & = 1 - \frac{1}{\ln\vert\mathcal{W}\vert}
         \left(\displaystyle\sum_{\substack{w,w'\in \mathcal{W}\\
                    a \in \mathcal{A}}}p(w',w,a)
                \ln\left[\frac{p(w'|w,a)}{p(w'|w)}\right]\right).
  \end{align*}
\end{mydef}

\begin{mydef}[Morphological Computation as positive effect of the world]
  \label{def:MC_W}
  Let the random variables $A,W,W'$ denote the action, the current and the next
  world state of an embodied agent, which is described by the causal diagram
  shown in Figure~\ref{fig:sml full causal}. The quantification of the
  morphological computation as the positive effect of the world on the
  behaviour is then defined as
  \begin{align*}
    \MCW & := \frac{1}{\ln\vert\mathcal{W}\vert} D(p(w'|w,a)||p(w'|a))\\
         & = \frac{1}{\ln\vert\mathcal{W}\vert}
         \left(\sum_{\substack{w,w'\in \mathcal{W}\\
           a \in \mathcal{A}}}p(w',w,a)
       \ln\left[\frac{p(w'|w,a)}{p(w'|a)}\right]\right).
  \end{align*}
\end{mydef}

The Kullback-Leibler divergence used in the first measure
$\mathrm{MC}_\mathrm{A}$ can also be rewritten as the 
conditional mutual information $I(W';A|W)$, which is related to the concept of
empowerment~\citep{Klyubin2005All-Else-Being}. The latter is defined as $C(w)=\max_{p(a)}I(W';A|w)$ and denotes how
empowered an embodied agent is in a specific state of the environment.
Morphological computation is correlated with a low influence 
the influence of the action $A$ on the next world state $W'$, whereas
empowerment requires the maximisation of the influence of the action on the next
world state.
Maximising morphological computation therefore minimises the
empowerment and vice versa. One may speculate that both concepts could be used
as balancing forces to find the optimal amount of control for an embodied
system.

Both definitions given above require full access to the world states $W$ and
$W'$. This is undesired for two reasons. First, it limits the applicability of
the measure to systems of low complexity which live in the domain of simple
grid-world environments. This contradicts our interest in presenting measures
that can be used to analyse natural or non-trivial artificial cognitive systems.
Second, we believe that the measures derived here may model an intrinsic driving
force in the context of guided self-organisation of embodied systems. Hence, we
require that any measure must operate on intrinsically available information
only. The resulting intrinsic measures are presented below, in the order of the
concepts they relate to. The next sections lists the four measures without
discussing them in detail. They are evaluated and discussed in the subsequent
sections.

%%%%%%%%%%%%%%%%%%%%%%%%%%%%%%%%%%%%%%%%%%%%%%%%%%%%%%%%%%%%%%%%%%%%%%%%%%%%%%%%
%%%%%%%%%%%%%%%%%%%%%%%%%%%%%%%%%%%%%%%%%%%%%%%%%%%%%%%%%%%%%%%%%%%%%%%%%%%%%%%%
%%%   ASOC_A
%%%%%%%%%%%%%%%%%%%%%%%%%%%%%%%%%%%%%%%%%%%%%%%%%%%%%%%%%%%%%%%%%%%%%%%%%%%%%%%%
%%%%%%%%%%%%%%%%%%%%%%%%%%%%%%%%%%%%%%%%%%%%%%%%%%%%%%%%%%%%%%%%%%%%%%%%%%%%%%%%

\subsubsection{Concept 1, Associative Measure}

The first measure that operates on intrinsic information only is the canonical
adaptation of the first definition to the intrinsic perspective. It must be
mentioned here, that the sensor states $S$ and $S'$ are understood as the
intrinsically available information about the external world. Hence, the
conditional probability $p(s'|s)$ refers to the intrinsically available
information about $p(w'|w)$. 

\begin{mydef}[Associative measure of the negative effect of the action]
  \label{def:ASOC_A}
  Let the random variables $A,S,S'$ denote the action, the current and the next
  sensor state of an embodied agent, which is described by the causal diagram
  shown in Figure~\ref{fig:sml full causal}. The quantification of the
  morphological computation as an associative measure of the negative effect of
  the action is then defined as:
  \begin{align}
    \ASOCA & := 1 -\frac{1}{\ln|\mathcal{S}|} D(p(s'|s,a)||p(s'|s)) \\
         & = 1 -\frac{1}{\ln|\mathcal{S}|}
         \sum_{\substack{s,s'\in \mathcal{S}\\ a \in \mathcal{A}}}
    p(s',s,a)\ln\left[\frac{p(s'|s,a)}{p(s'|s)}\right]
  \end{align}
\end{mydef}
The Kullback-Leibler divergence used in the definition of \ASOCA (see
Def.~\ref{def:ASOC_A}) is also know as the transfer entropy of $A$ on $S$
\citep{Schreiber2000Measuring-Information-Transfer} and it was investigated in
the context to quantify the informational structure of sensory and
motor data of embodied agents \citep{Lungarella2005Methods-for-quantifying}.

%%%%%%%%%%%%%%%%%%%%%%%%%%%%%%%%%%%%%%%%%%%%%%%%%%%%%%%%%%%%%%%%%%%%%%%%%%%%%%%%
%%%%%%%%%%%%%%%%%%%%%%%%%%%%%%%%%%%%%%%%%%%%%%%%%%%%%%%%%%%%%%%%%%%%%%%%%%%%%%%%
%%%   CAUSAL A
%%%%%%%%%%%%%%%%%%%%%%%%%%%%%%%%%%%%%%%%%%%%%%%%%%%%%%%%%%%%%%%%%%%%%%%%%%%%%%%%
%%%%%%%%%%%%%%%%%%%%%%%%%%%%%%%%%%%%%%%%%%%%%%%%%%%%%%%%%%%%%%%%%%%%%%%%%%%%%%%%

\subsubsection{Concept 1, Causal Measure}

\begin{figure}[t]
  \begin{center}
    \subfigure[Schematics of the sensori-motor loop for a reactive system]{
        \includegraphics[height=4.25cm]{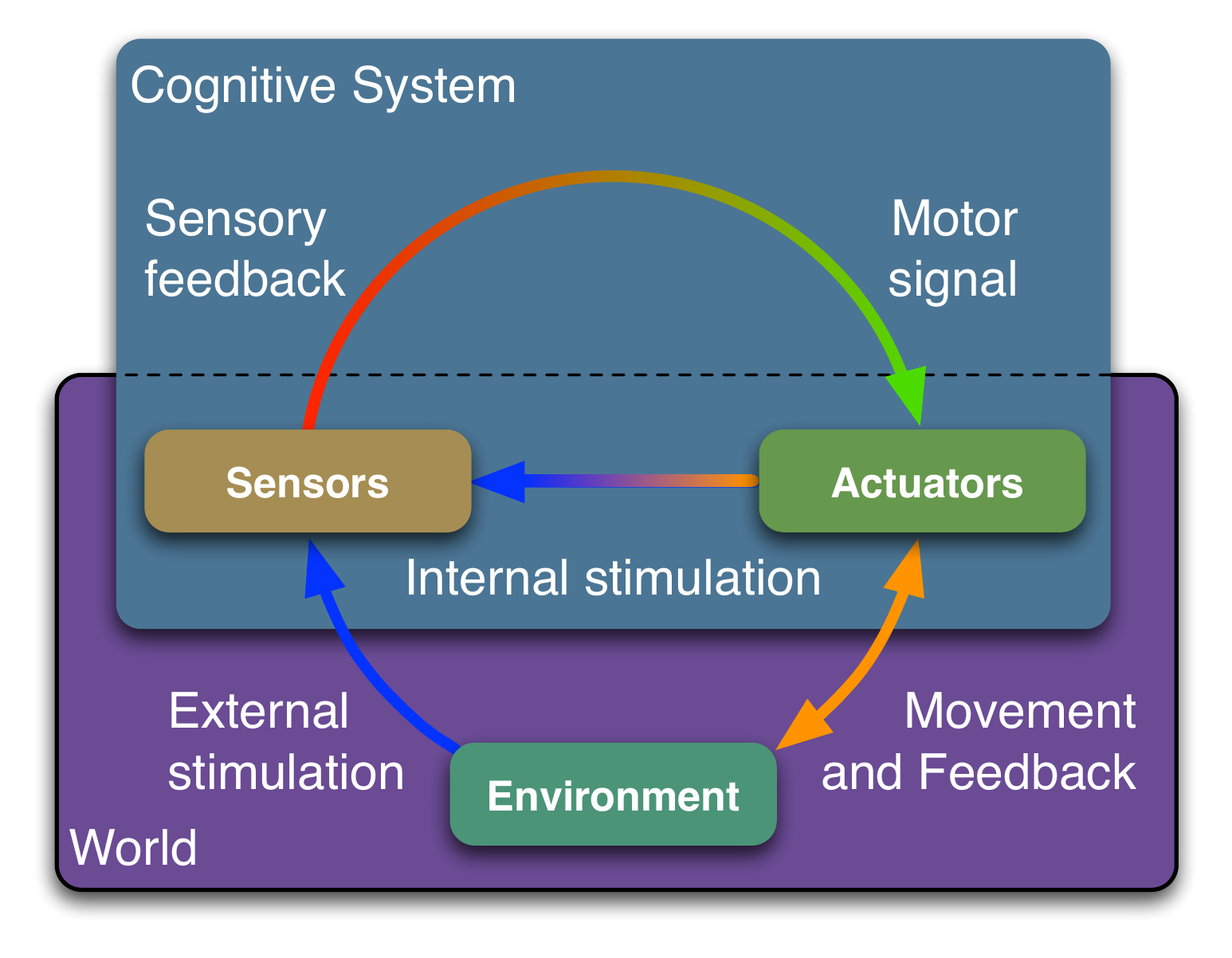}\label{fig:sml reactive concept}}
    \hspace*{2em}
    \subfigure[Representation of the sensori-motor loop for a reactive system as a causal graph]{
      \includegraphics[height=4.25cm]{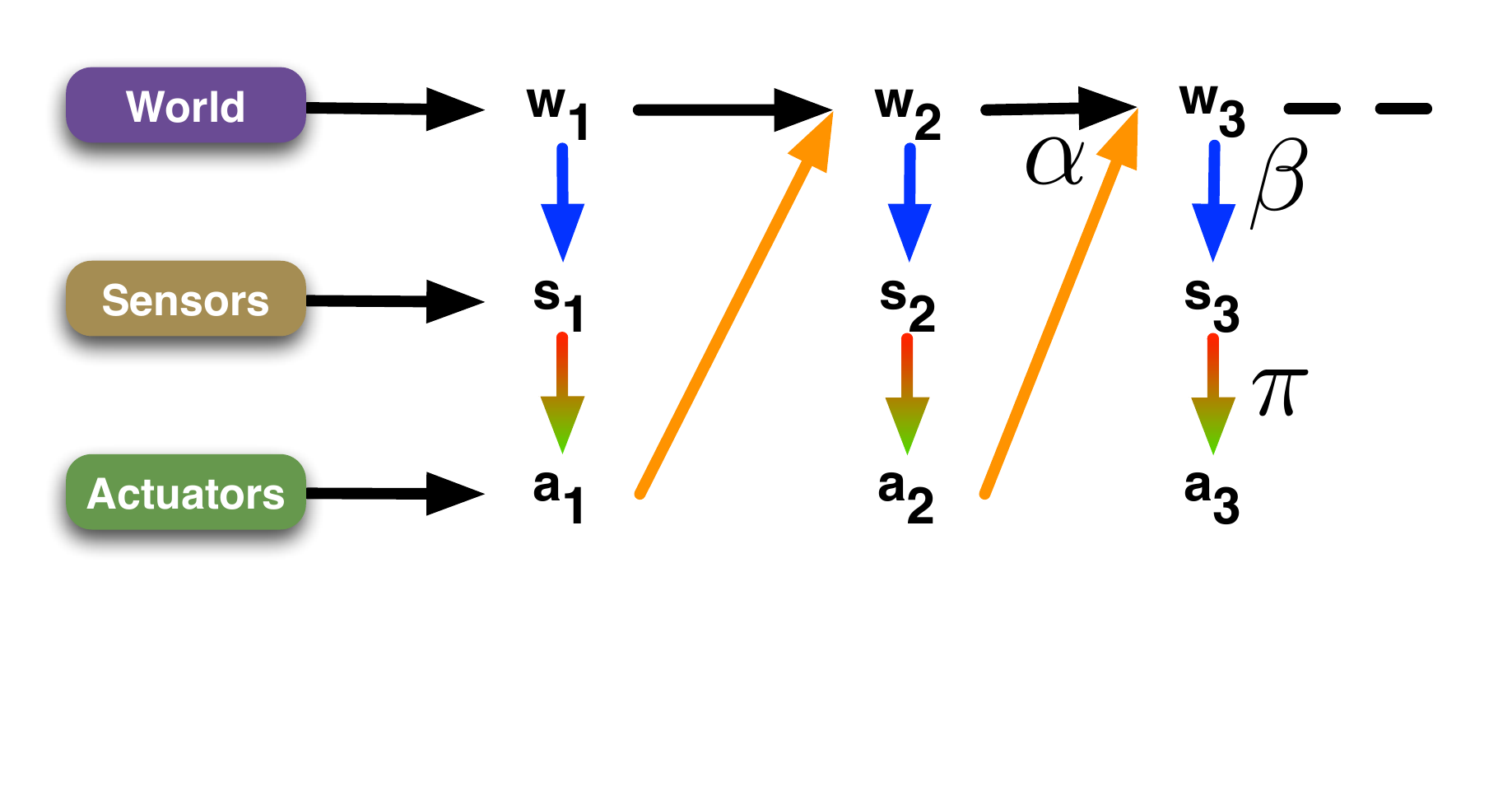}\label{fig:sml reactive causal}}
      \caption{Reactive system. In the context of this work, a reactive system
        is defined by a direct coupling of the sensors and actuators.
        There is no form of memory present in the system.}
    \label{fig:sml reactive}
  \end{center}

  % \begin{center}
  %   \includegraphics[height=3.25cm]{sensori-motor-loop3.pdf}
  % \end{center}
  % \caption{Causal graph of the sensori-motor loop for a reactive system and one
  %   time-step only. A reactive controller does not have any internal controller
  %   state $C$. Therefore, the entire history of the system is fully determined
  %   by the last world state $W$. The sensors are directly connected to the }
  % \label{fig:reduced reactive sml}
\end{figure}

The second measure which follows the first concept is based on measuring the
causal information flow of the sensor and action states $S,A$ on the next sensor
state $S'$. To simplify the argumentation, we first consider a reactive
system (see Fig.~\ref{fig:sml reactive causal}). The measure will then also be presented
and discussed for non-reactive systems, as both cases result in different
measures. The idea for this measure is captured in the Figure~\ref{fig:causal
idea}. Please keep in mind, that the sensor states $S$ and $S'$ are here
understood as internally available information about the system's \emph{Umwelt}.
The causal information flow from $S$ to $S'$, denoted by
$CIF(S\rightarrow S')$, includes the information that flows from $S$ to $S'$ over
all pathways. This explicitly includes the information flow from $S$ to $S'$
over the action $A$ (see Fig.~\ref{fig:causal idea s,s'}). Morphological
computation is here understood as the causal information flow from $S$ to $S'$
excluding the causal information flowing from the action $A$ to $S'$ (see
Fig.~\ref{fig:causal idea s,s',a}). Hence, we need to subtract the causal
information flow from $A$ to $S'$, denoted by $CIF(A\rightarrow S')$ (see
Fig.~\ref{fig:causal idea a,s'}) to exclude it from $CIF(S\rightarrow S')$ such
that we receive the causal information flow that goes from $S$ to $S'$ without passing
$A$, denoted by $CIF(S\rightarrow S' \backslash A)$ (see Fig.~\ref{fig:causal
idea s,s',a}).

\begin{figure}[t]
  \begin{center}
    \subfigure[$CIF(S\rightarrow S')$]{\label{fig:causal idea s,s'}
     \includegraphics[height=3cm]{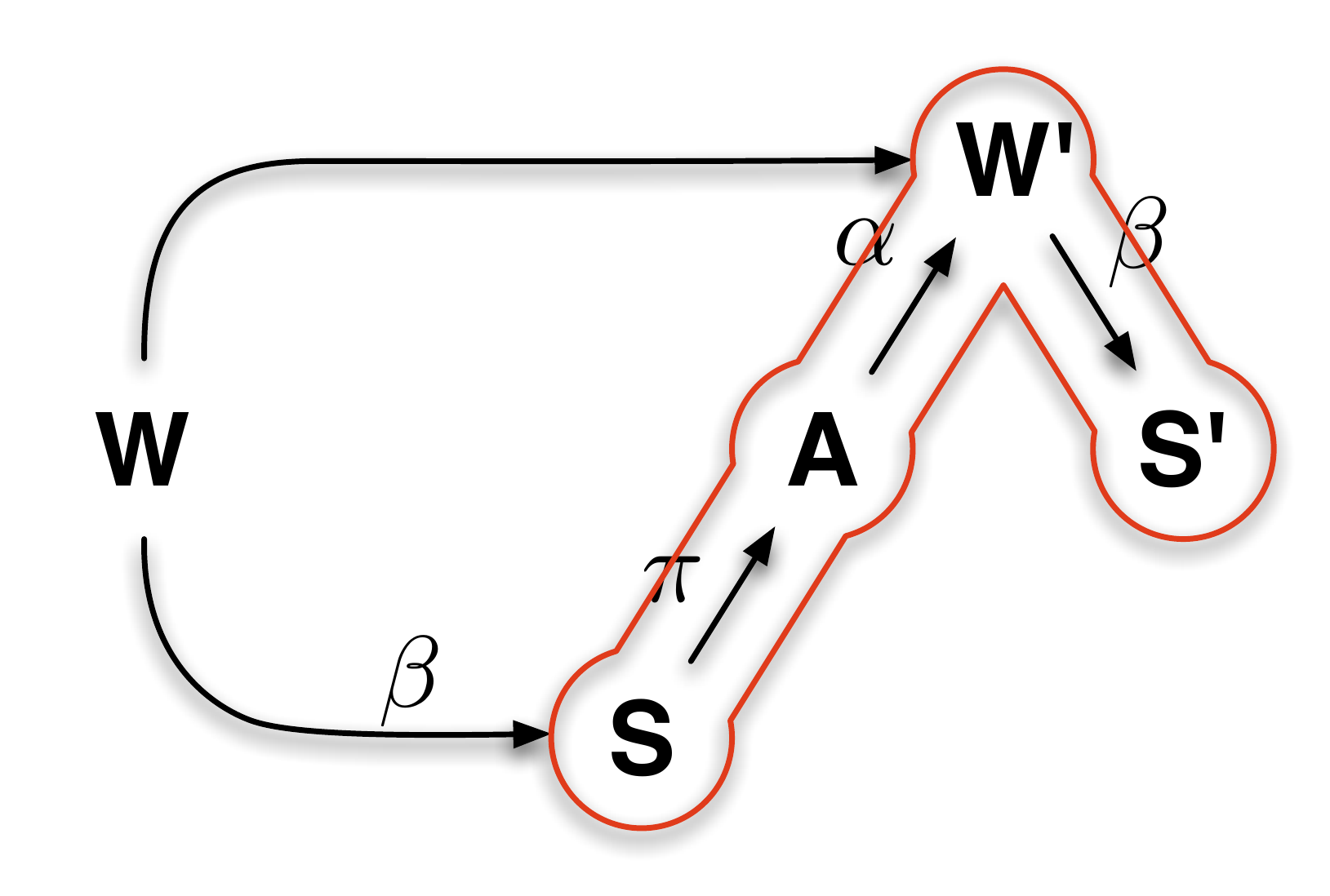}}
    \subfigure[$CIF(A\rightarrow S')$]{\label{fig:causal idea a,s'}
     \includegraphics[height=3cm]{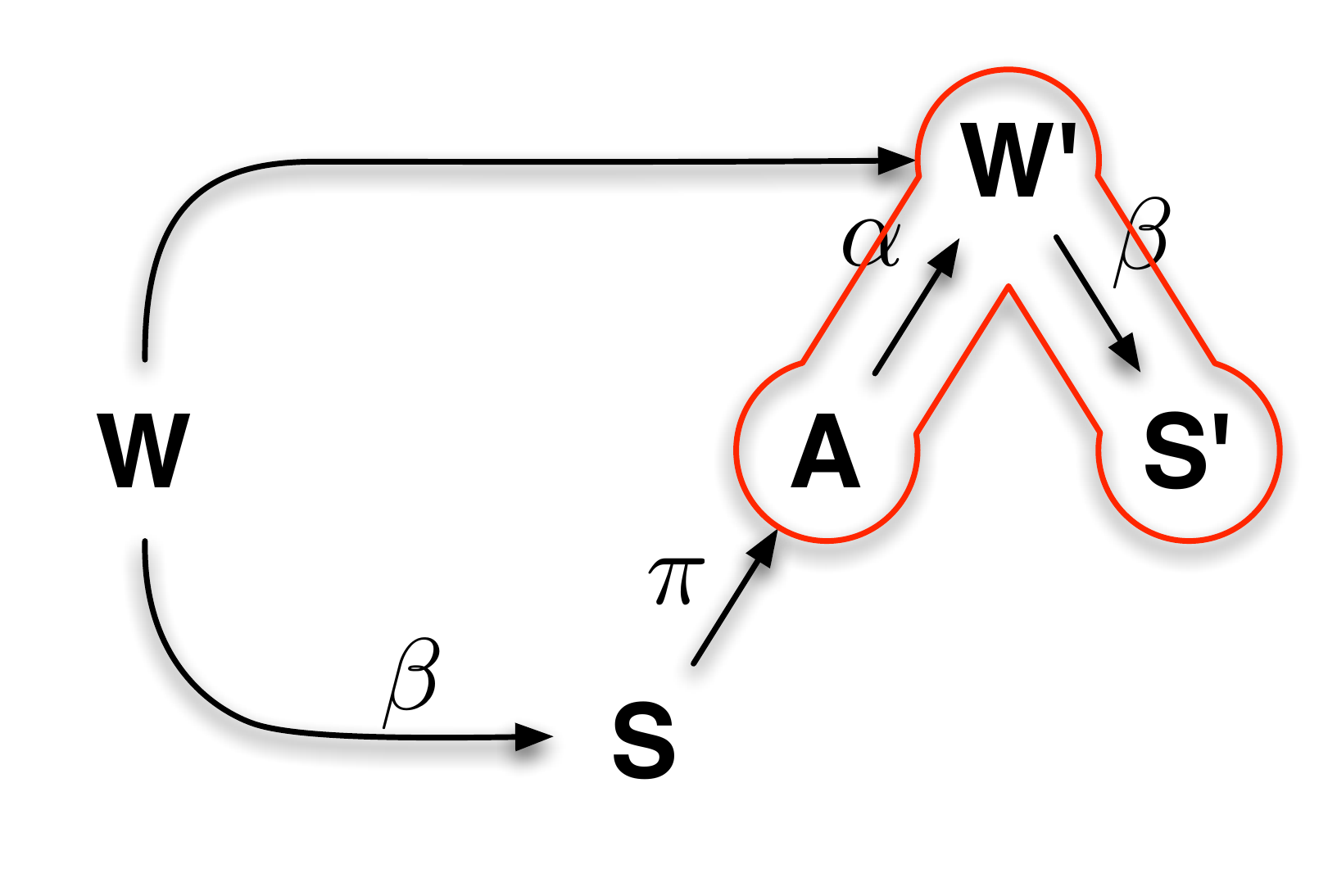}}
    \subfigure[$CIF(S\rightarrow S' \backslash A)$]{\label{fig:causal idea s,s',a}
     \includegraphics[height=3cm]{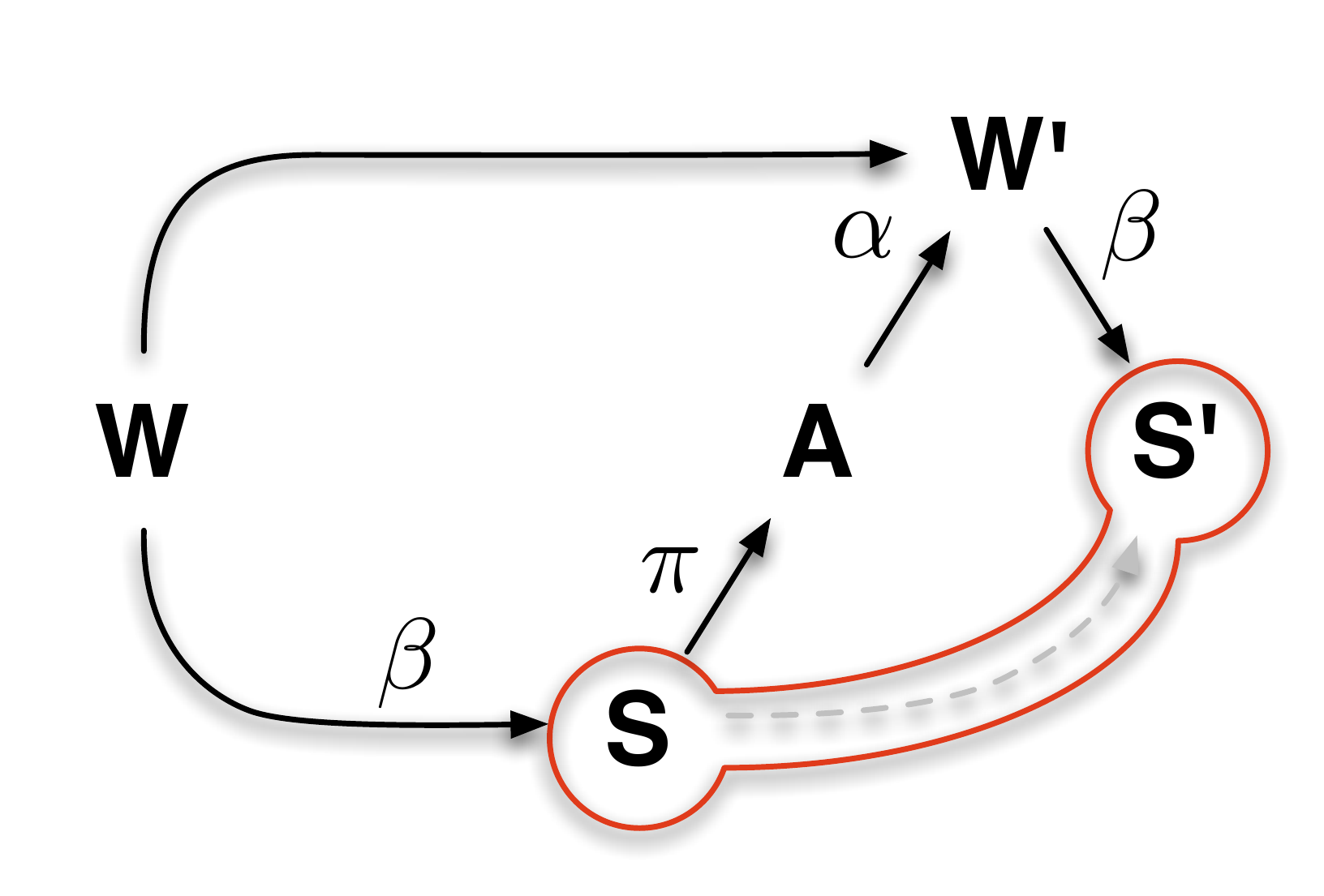}}
  \end{center}
  \caption{Visualisation of the causal measure \CA. The causal graph used
    in the Figures (a), (b), and (c)
    is the reduction of
    sensori-motor loop shown in Figure~\ref{fig:sml reactive causal} to
    two consecutive time steps.
    Figure (a) shows that the
    causal information flow $CIF(S\rightarrow S')$ measures all causal
    information from $S$ to $S'$, including the information that flows over $A$.
    Figure (b) shows that the causal information flow $CIF(A\rightarrow S')$
    only captures the information flowing from $A$ to $S'$. Both can be used to
    approximate the causal information from $S$ to $S'$, that does not pass
    through $A$, denoted by $CIF(S\rightarrow S'\backslash A)$, as shown Figure
    (c).}
  \label{fig:causal idea}
\end{figure}

According to the general theory \cite{Pearl2000Causality:-Models-Reasoning},
we talk about identifiable causal effects if they are computable from
observational data.
We stated earlier, that we are interested in intrinsic
measures. Hence, we require that the causal effects are identifiable from
observational data that are intrinsically available to the agent. We refer to
this type of identifiability as \emph{intrinsically identifiable}.
In our previous work \citep{Ay2013An-Information-Theoretic}, we
showed that the causal effects of $S$ on $S'$ and of $A$
on $S'$ are intrinsically identifiable by the following conditional probability
distributions:
\begin{align}
  p(s'|\mathrm{do}(a))   & =
  \sum_{s\in\mathcal{S}}p(s'|s,a)p(s)\label{eq:p(s'|do(a))}\\
  p(s'|\mathrm{do}(s))   & =
  \sum_{a\in\mathcal{A}}p(a|s)\sum_{s''\in\mathcal{S}}p(s'|s'',a)p(s'')\label{eq:p(s'|do(s))}\\
  & = \sum_{a\in\mathcal{A}}p(a|s) p(s'|\mathrm{do}(a)) \label{eq:equivalence}
\end{align}
These distributions need to be explained. The notation $p(x|\mathrm{do}(y))$
(also denoted by $p(x|\hat{y})$ refers to probability of measuring $x$ when the
state of $Y$ was set by intervention to $y$
\citep{Pearl2000Causality:-Models-Reasoning}. Explicitly, this means that
intervention is generally required in order to determine causation. Therefore, it
is important to note, that the equations above (see Eq.~\ref{eq:p(s'|do(a))} and
\ref{eq:p(s'|do(s))}) allow us to determine the causal effects \emph{without}
any intervention. An agent can act in the sensori-motor loop, and from its
observation determine e.g.~the causal effect of its actions $A$ on the
its next sensor states $S'$ \citep[for a discussion,
see][]{Ay2013An-Information-Theoretic}. From the two probability
distributions given in the Equations~\ref{eq:p(s'|do(a))} and
\ref{eq:p(s'|do(s))} we can construct the two required causal information
measures for $CIF(S\rightarrow S')$ and $CIF(A\rightarrow S')$
\citep[see also][]{Ay2008Information-Flows-in}. The derivations
for both measures are given in the appendix (see Sec.~\ref{app:deviation of
causal measure 1}).
The resulting equations are given by
\begin{align}
  CIF(S\rightarrow S')  & = \sum_{s\in\mathcal{S}}p(s)\sum_{s'\in\mathcal{S}} 
  p(s'|\mathrm{do}(s)) \ln
  \left[\frac{p(s'|\mathrm{do}(s))}
  {\sum_{s''\in\mathcal{S}}p(s'|\mathrm{do}(s''))p(s'')} \right]\\
  CIF(A\rightarrow S') & = \sum_{a\in\mathcal{A}}p(a)\sum_{s'\in\mathcal{S}} 
  p(s'|\mathrm{do}(a)) \ln
  \left[\frac{p(s'|\mathrm{do}(a))}
  {\sum_{a'\in\mathcal{A}}p(s'|\mathrm{do}(a'))p(a')}
  \right]
\end{align}
The difference $CIF(S\rightarrow S') - CIF(A\rightarrow S')$ is always negative
(see Eq.~\ref{eq:CS smaller CA} in Sec.~\ref{app:deviation of causal measure
1}), and hence, the resulting
measure is given by the following definition.
\begin{mydef}[Causal measure of the negative effect of the action for a reactive system]
  \label{def:CAUSAL_A_1}
  Let the random variables $A,S,S'$ denote the action, the current and the next
  sensor state of a reactive embodied agent, which is described by the causal diagram
  shown in Figure~\ref{fig:sml full causal}. The quantification of the
  morphological computation as a causal measure of the negative effect of
  the action is then defined as:
  \begin{align}
    \CA & := 1 + \frac{1}{\ln|\mathcal{S}|}(CIF(S\rightarrow S') -
    CIF(A\rightarrow S')) \label{eq:causal bottleneck}\\
        &  = 1 -
        \frac{1}{\ln|\mathcal{S}|}D(p(s'|\mathrm{do}(a))||p(s'|\mathrm{do}(s)))
        \label{eq:D do a do s}\\
        &  = 1 - \frac{1}{\ln|\mathcal{S}|}\sum_{s\in\mathcal{S},a\in\mathcal{A}}p(s,a)\sum_{s'\in\mathcal{S}}p(s'|\mathrm{do}(a))\ln\frac{p(s'|\mathrm{do}(a))}{p(s'|\mathrm{do}(s))}
  \end{align}
\end{mydef}

The causal information flow $CIF(S\rightarrow S')$ is so far not shown to be
intrinsically identifiable for non-reactive systems
\cite{Ay2013An-Information-Theoretic}. Therefore, we consider the
causal information flow from the internal controller state $C$ to the next
sensor state $S'$ (see Fig.~\ref{fig:sml full causal}). This is valid, because $C$
represents the entire history of the system, and therefore, also the internal
representation of the entire history of the world. All further calculations are
analogous to the previous case \citep[see also][for a
discussion]{Ay2013An-Information-Theoretic} and lead to the following
definition.
\begin{mydef}[Causal measure of the negative effect of the action for a
  non-reactive system]
  \label{def:CAUSAL_A_2}
  Let the random variables $A,C,S'$ denote the action, the controller and the next
  sensor state of a non-reactive or deliberative embodied agent, which is
  described by the causal diagram shown in Figure~\ref{fig:sml full causal}. The
  quantification of the morphological computation as a causal measure of the
  negative effect of the action is then defined as:
  \begin{align}
  \mathrm{C}^\mathrm{d}_\mathrm{A} 
    & := 1 + \frac{1}{\ln|\mathcal{S}|}(CIF(C\rightarrow S') - CIF(A\rightarrow S')) \\
        &  = 1 - \frac{1}{\ln|\mathcal{S}|}D(p(s'|\mathrm{do}(a))||p(s'|\mathrm{do}(c)))\\
        &  = 1 - \frac{1}{\ln|\mathcal{S}|}\sum_{c\in\mathcal{C},a\in\mathcal{A}}p(c,a)\sum_{s'\in\mathcal{S}}p(s'|\mathrm{do}(a))\ln\frac{p(s'|\mathrm{do}(a))}{p(s'|\mathrm{do}(c))}
  \end{align}
\end{mydef}

The last definition (see Def.~\ref{def:CAUSAL_A_2}) is given for the
reason of completeness only. Morphological computation is mainly discussed in the
context of behaviours which are well-modelled as reactive behaviours
(e.g.~locomotion). To the best of our knowledge, it has so far not been
discussed in the context of non-reactive or
deliberative behaviours. Therefore, the first definition of the causal measure
(see Def.~\ref{def:CAUSAL_A_1}) suffices for all currently discussed cases of
morphological computation in the field of embodied artificial intelligence.

This concludes the measures of the first concept, in which morphological
computation is calculated inversely proportional to the influence the action $A$
has on the next world
state $W'$. The next section discusses the measures for morphological
computation in which it is calculated proportionally to the effect the world has
on itself.

%%%%%%%%%%%%%%%%%%%%%%%%%%%%%%%%%%%%%%%%%%%%%%%%%%%%%%%%%%%%%%%%%%%%%%%%%%%%%%%%
%%%%%%%%%%%%%%%%%%%%%%%%%%%%%%%%%%%%%%%%%%%%%%%%%%%%%%%%%%%%%%%%%%%%%%%%%%%%%%%%
%%%   ASOC W
%%%%%%%%%%%%%%%%%%%%%%%%%%%%%%%%%%%%%%%%%%%%%%%%%%%%%%%%%%%%%%%%%%%%%%%%%%%%%%%%
%%%%%%%%%%%%%%%%%%%%%%%%%%%%%%%%%%%%%%%%%%%%%%%%%%%%%%%%%%%%%%%%%%%%%%%%%%%%%%%%

\subsubsection{Concept 2, Associative Measure}

Analogous to the associative measure of the first concept, the associative
measure of the second concept adapts the Definition~\ref{def:MC_W} to the
intrinsic perspective by replacing $W$ by $S$ and $W'$ by $S'$. This leads to
the following definition.
\begin{mydef}[Associative measure of the positive effect of the world]
  \label{def:ASOC_W}
  Let the random variables $A,S,S'$ denote the action, the current and the next
  sensor state of an embodied agent, which is described by the causal diagram
  shown in Figure~\ref{fig:sml full causal}. The quantification of the
  morphological computation as an associative measure of the positive effect of
  the world is then defined as:
  \begin{align}
    \ASOCW & := \frac{1}{\ln|\mathcal{S}|}D(p(s'|s,a)||p(s'|a)) \\
        & = \frac{1}{\ln|\mathcal{S}|}
        \sum_{\substack{s,s'\in \mathcal{S}\\ a \in \mathcal{A}}}
        p(s',s,a)\ln\left[\frac{p(s'|s,a)}{p(s'|a)}\right].
  \end{align}
\end{mydef}

The next measure approximates the dependence of the next world state $W'$ on the
current world state $W$ solely based on the internal world model $p(s'|s,a)$.

%%%%%%%%%%%%%%%%%%%%%%%%%%%%%%%%%%%%%%%%%%%%%%%%%%%%%%%%%%%%%%%%%%%%%%%%%%%%%%%%
%%%%%%%%%%%%%%%%%%%%%%%%%%%%%%%%%%%%%%%%%%%%%%%%%%%%%%%%%%%%%%%%%%%%%%%%%%%%%%%%
%%%   CONDITIONAL INDEPENDENCE
%%%%%%%%%%%%%%%%%%%%%%%%%%%%%%%%%%%%%%%%%%%%%%%%%%%%%%%%%%%%%%%%%%%%%%%%%%%%%%%%
%%%%%%%%%%%%%%%%%%%%%%%%%%%%%%%%%%%%%%%%%%%%%%%%%%%%%%%%%%%%%%%%%%%%%%%%%%%%%%%%

\subsubsection{Concept 2, Conditional Independence}

\begin{figure}[h]
  \begin{center}
    \includegraphics[height=4cm]{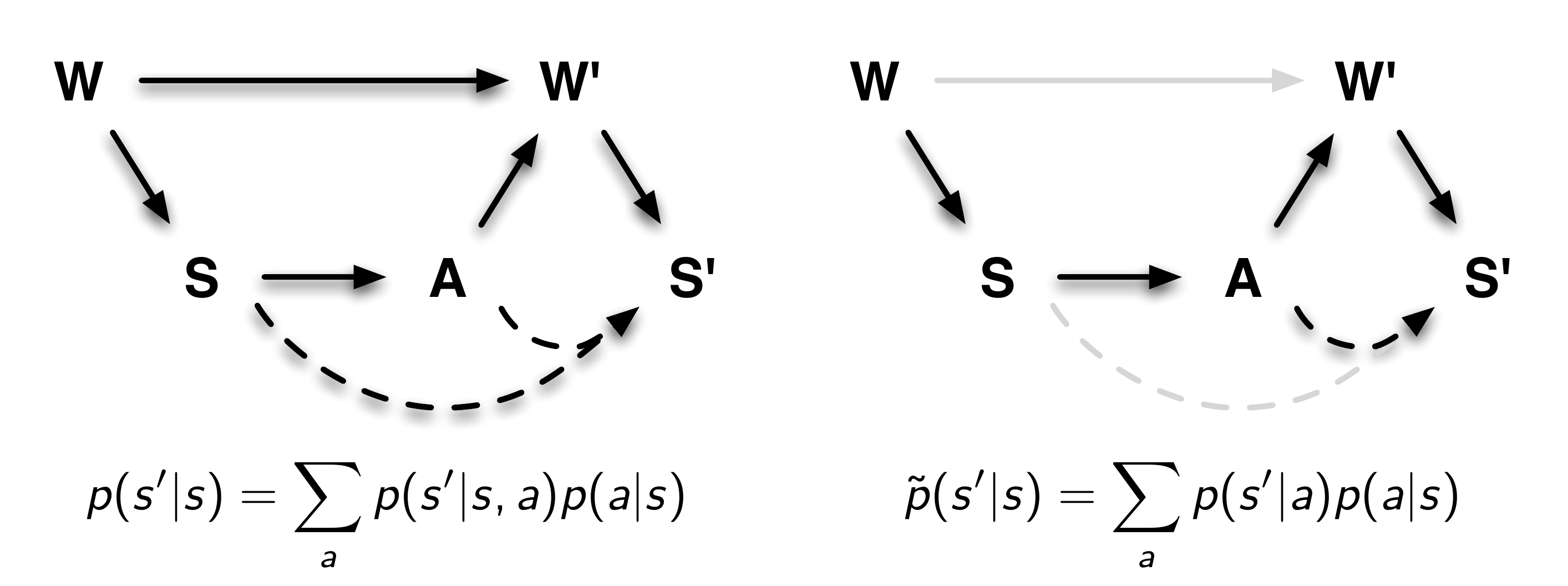}
  \end{center}
  \caption{Visualisation of the conditional independence measure \CW. The left-hand
    side shows how the conditional probability distributions $p(s'|s)$ can be
    calculated from the world model $p(s'|s,a)$ and the policy $p(a|s)$. The
    right-hand side shows how $p(s'|s)$ changes, if one assumes that the world
    does not influence itself (gray arrow between $W$ and $W'$), and if this is
    reflected in the internal world model (gray arrow between $S$ and $S'$). The
    difference of both measures the morphological computation.}
  \label{fig:m2 prime concept}
\end{figure}
The idea of this measure is to calculate how much the conditional probability
distribution $p(s'|s)$ estimated from the recoded data differs from the
assumption that the world did not have any effect on itself, denoted by the
conditional probability distribution $\tilde{p}(s'|s)$. The Figure~\ref{fig:m2
prime concept} shows the difference between the two conditional probability
distributions graphically. The following equations show how they can be
calculated from the observed data.
\begin{align}
  p(s'|s) & = \sum_{a\in\mathcal{A}}p(s'|s,a) p(a|s)
  \label{eq:p(s'|s) concept 2}\\
\tilde{p}(s'|s) & = \sum_{a\in\mathcal{A}}p(s'|a)p(a|s) =
  \sum_{a\in\mathcal{A}}p(a|s)
  \sum_{s''\in\mathcal{S}}p(s'|s'',a)p(a|s'') \frac{p(s'')}{p(a)}
  \label{eq:tildep(s'|s) concept 2}
\end{align}
The measure is then defined as the Kullback-Leibler divergence of $p(s'|s)$ and
$\tilde{p}(s'|s)$.

\begin{mydef}[Conditional dependence of the world on itself]
  \label{def:C_W}
  Let the random variables $A,S,S'$ denote the action, the current and the next
  sensor state of an embodied agent, which is described by the causal diagram
  shown in Figure~\ref{fig:sml full causal}. The quantification of the
  morphological computation as the error of the assumption, that the next
  world state is conditionally independent of the previous world state 
  is then defined as:
  \begin{align}
    \CW & := \frac{1}{\ln|\mathcal{S}|}D(p(s'|s)||\hat{p}(s'|s))\\
    & = \frac{1}{\ln|\mathcal{S}|}
    \sum_{s\in\mathcal{S},s'\in\mathcal{S}}
    p(s'|s)p(s) \ln \left[\frac{p(s'|s)}{\tilde{p}(s'|s)}\right]
  \end{align}
\end{mydef}

This concludes the presentation of definitions of morphological computation
measures. The next section discusses their relation to the Information
Bottleneck Method \cite{Tishby1999The-information-bottleneck}.

\subsection{Relation to the Information Bottleneck Method}
\begin{figure}[h]
  \begin{center}
    \subfigure[Concept of morphological computation\label{fig:ibm1}]{
      \includegraphics[height=3cm]{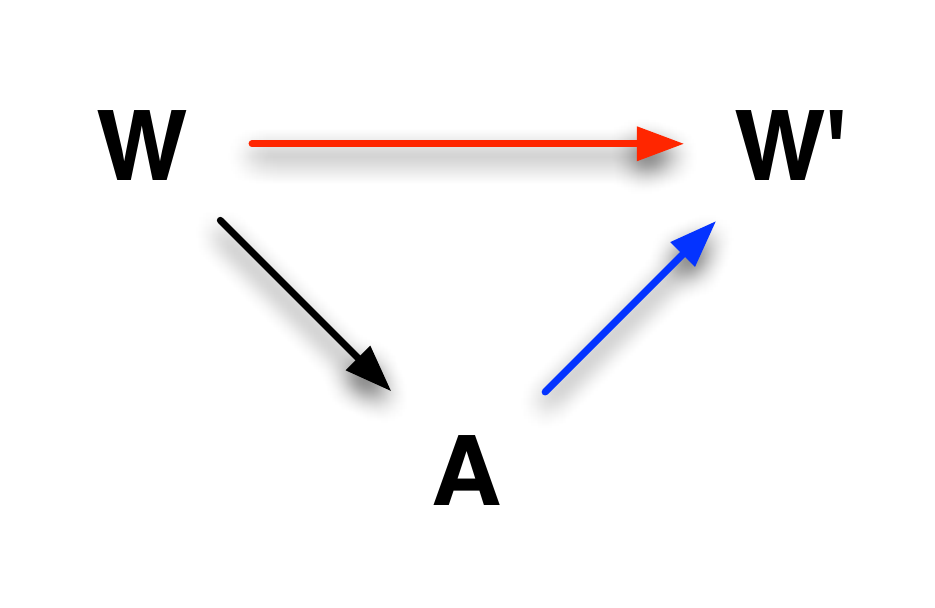}}
    \hspace*{3ex}
    \subfigure[Concept of the Information Bottleneck
    Method\label{fig:ibm2}]{\includegraphics[height=3cm]{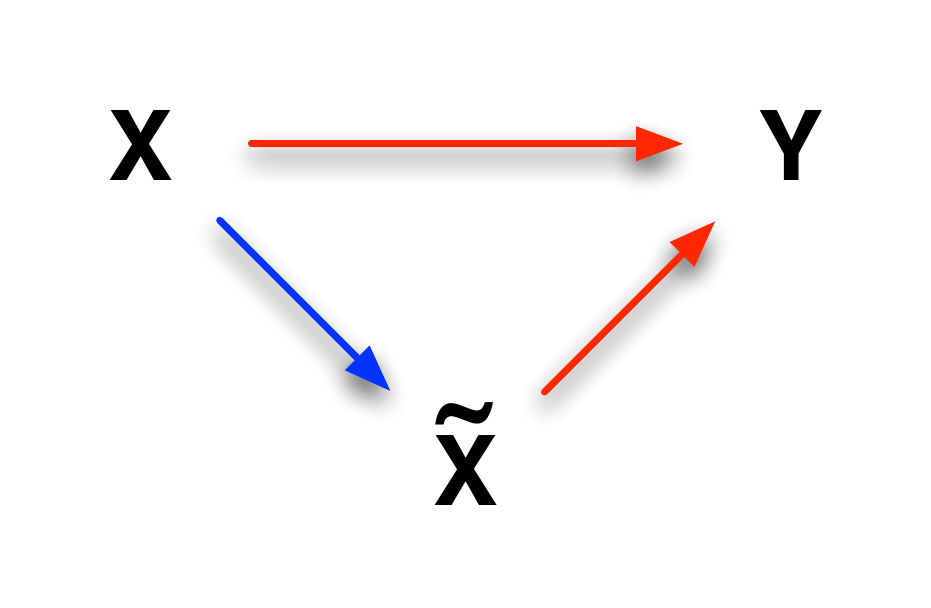}}
  \end{center}
  \caption{Visualisation of the relation of the concepts to the Information
    Bottleneck Method. Figure (a) shows the general concept of morphological
    computation. The world states are highly correlated (red arrow), whereas the
    world and action are only weakly correlated (blue arrow). The Figure (b)
    show the concept of the Information Bottleneck Method. The difference is
    that a strong correlation between the action $A$ and $W'$ and a weak
    correlation between $W$ and $A$ are required. For a discussion, please read
    the text below.}
  \label{fig:ibm}
\end{figure}

We will discuss the relation of morphological computation to the Information
Bottleneck Method by \citet{Tishby1999The-information-bottleneck} along with the
two graphs shown in Figure~\ref{fig:ibm}. The graph on the left-hand side (see
Fig.~\ref{fig:ibm1}) shows how we understand morphological computation. It is
present, if the current world state $W$ has a high influence on the next world
state $W'$ (denoted by a red arrow) and the current action $A$ has a low
influence on the next world state $W'$ (denoted by a blue arrow). Hereby, it is
not relevant how much the world has influenced the action (denoted by a black
arrow).

The Information Bottleneck Method is visualised in the Figure~\ref{fig:ibm2}.
Here, the influence of the world on the action is minimised and the influence of
the action on the next world state is maximised. The action $A$ is the
bottleneck variable $\tilde{X}$ (see Fig.~\ref{fig:ibm2}) for the relevant information
that is passed through the world. This appears connected for the following
reason. Conceptionally speaking, reducing the required alphabet in $A$ to still
maintain the information the world carries about itself is related to the notion
of morphological computation as it means that only a few actions should be
chosen by the agent to maintain the behaviour. Formally, the two concepts
(Information Bottleneck and morphological computation) are not easily aligned.
The Information Bottleneck Method requires to minimised the Kullback-Leibler
divergence $D(p(w'|w)||p(w'|a))$ \citep[see e.g.~Eq.~(27)
in][]{Tishby1999The-information-bottleneck}, whereas morphological computation
is defined to minimise the Kullback-Leibler divergence $D(p(w'|w,a)||p(w'|w))$
(see Def.~\ref{def:MC_A})
or to maximise the Kullback-Leibler divergence $D(p(w'|w,a)||p(w'|a))$  
(see Def.~\ref{def:MC_W}). We could currently not solve this contradiction, but we
believe that it should be followed, as it could unify the presented concepts
\MCA and \MCW.

This concludes the presentation and discussion of the two concepts, their
formalisations, and the resulting four measures that rely on internal variables
only. The next step in this work is to evaluate how these different measure
behave, when they are applied to experiments. This is done in the next section.

%%%%%%%%%%%%%%%%%%%%%%%%%%%%%%%%%%%%%%%%%%%%%%%%%%%%%%%%%%%%%%%%%%%%%%%%%%%%%%%%
%%%%%%%%%%%%%%%%%%%%%%%%%%%%%%%%%%%%%%%%%%%%%%%%%%%%%%%%%%%%%%%%%%%%%%%%%%%%%%%%
%%%   EXPERIMENTS
%%%%%%%%%%%%%%%%%%%%%%%%%%%%%%%%%%%%%%%%%%%%%%%%%%%%%%%%%%%%%%%%%%%%%%%%%%%%%%%%
%%%%%%%%%%%%%%%%%%%%%%%%%%%%%%%%%%%%%%%%%%%%%%%%%%%%%%%%%%%%%%%%%%%%%%%%%%%%%%%%

\section{Experiments}
\label{sec:experiments}

This section applies the morphological computation measures to two different
experiments. The first experiment is a simplified parameterisable model of the
one-step sensori-motor loop for reactive systems (see Fig.~\ref{fig:sml reactive
causal}).
The model is defined by the four transition maps
$\alpha_{\phi,\psi}(w'|w,a)$, $\beta_\zeta(s|w)$, $\pi_\mu(a|s)$, and
$p_\tau(w)$. For each map, the indices refer to its
parameters. Hence, the entire model is parameterised by five parameters, of
which two are kept constant in the experiments below.

The second experiment is designed as a minimal
physical system, that allows a transition between the Dynamic Walker
\citep{Wisse2004Essentials-of-dynamic} and a classically controlled humanoid.

Both experiments use reactive systems for the reason of
simplicity only. This does not mean that all measures are limited to reactive
systems. In particular, the two concepts $\mathrm{MC}_\mathrm{W}$ and
$\mathrm{MC}_\mathrm{A}$ and the intrinsic adaptations
$\mathrm{ASOC}_\mathrm{W}$, $\mathrm{ASOC}_\mathrm{A}$,
$\mathrm{C}^\mathrm{d}_\mathrm{A}$ and \CW make no assumptions
on the type of control. Only \CA, which is used in the following experiments,
explicitly requires a reactive control.

%%%%%%%%%%%%%%%%%%%%%%%%%%%%%%%%%%%%%%%%%%%%%%%%%%%%%%%%%%%%%%%%%%%%%%%%%%%%%%%%
%%%%%%%%%%%%%%%%%%%%%%%%%%%%%%%%%%%%%%%%%%%%%%%%%%%%%%%%%%%%%%%%%%%%%%%%%%%%%%%%
%%%   BINARY MODEL
%%%%%%%%%%%%%%%%%%%%%%%%%%%%%%%%%%%%%%%%%%%%%%%%%%%%%%%%%%%%%%%%%%%%%%%%%%%%%%%%
%%%%%%%%%%%%%%%%%%%%%%%%%%%%%%%%%%%%%%%%%%%%%%%%%%%%%%%%%%%%%%%%%%%%%%%%%%%%%%%%

\subsection{Binary Model Experiment}
\label{sec:binary model}

In the introduction to this paper, it was stated that the measures presented in
this work should be applicable to biological systems. In the previous section
(see Sec.~\ref{sec:measuring morphological computation}), it was stated that the
measures should not operate on the world state $W$, also because it is not
accessible for any example other then simple toy worlds. This section now
evaluates the two concepts and four measures based on such a simple toy world
example. This might appear as a contradiction to the previous statements, and
therefore, the application of the measures to this model needs a further
explanation.

In order to understand how the measures differ and in which aspects they are
alike, it is best to analyse them under a fully controllable setting. Using real
robots only allows to fully control a very limited set of parameters of the
entire system, namely the parameters of the policy and partially those of the
embodiment. This improves only slightly in simulated robotics, depending on the
chosen implementation. These uncontrollable or hidden parameters require
robustness of the applied methods, and are one of the main arguments 
for favouring virtual or real robots over grid world environments. The goal of this
section is different. Here, we want to validate the measures by controlling the
influence of the world on itself in addition to controlling the systems
parameters. Therefore, it is necessary to
choose an experiment which allows us to fully control every aspect of the
sensori-motor loop, explicitly including the world transition kernel
$\alpha(w'|a,w)$. The next section will then validate the measures in a more
realistic experiment.

\begin{figure}[th]
  \begin{center}
    \includegraphics[height=3cm]{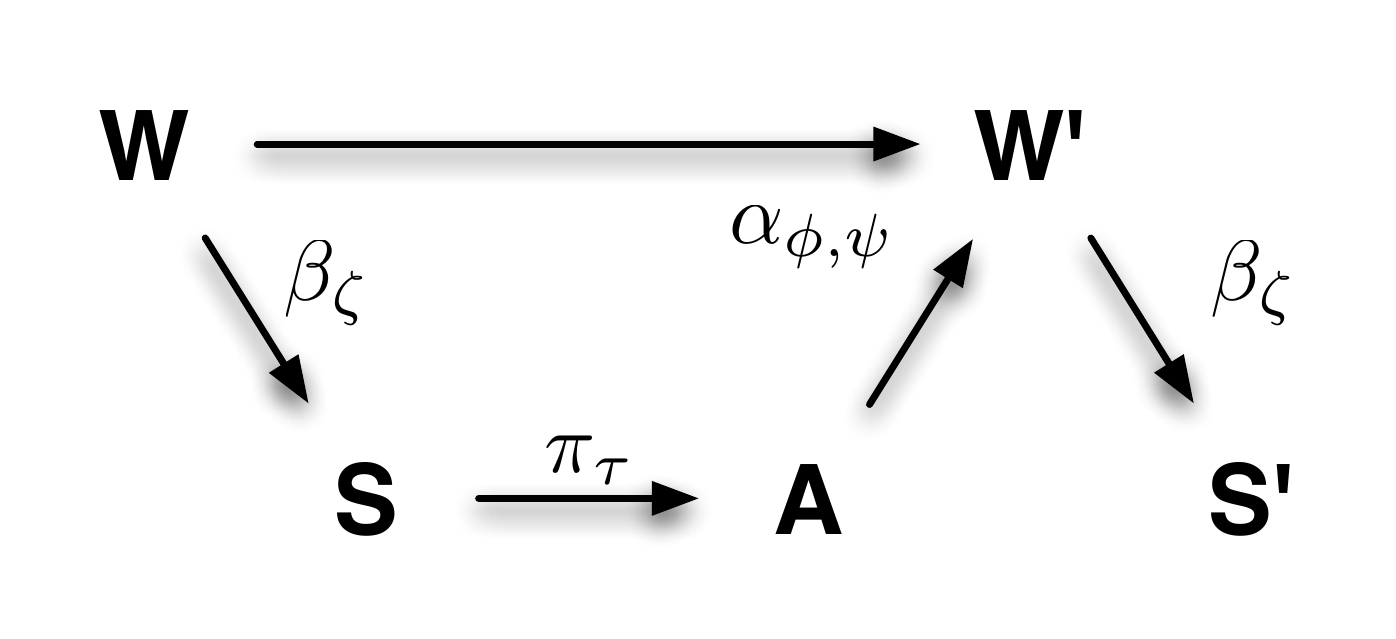}
  \end{center}
  \caption{Visualisation of the binary model. This figure shows the causal graph
    that is used in the binary model experiment (see text below).
    It is the graph representing the single step sensori-motor loop of a
    reactive system, where the indices of the transition maps refer to their
    parametrisation.}
  \label{fig:binary model}
\end{figure}

The minimalistic model discussed in this section is shown in Figure~\ref{fig:binary model} and is defined by the following set of equations:\\
\begin{minipage}[t]{0.5\textwidth}
  {\begin{align}
    \alpha_{\phi,\psi}(w'|w,a) & = \frac{e^{\phi w'w + \psi w'a}}%
                            {\sum_{w''\in\Omega} e^{\phi w''w + \psi w''a}}
                            \label{eq:p(w'|w,a)}\\
    \beta_\zeta(s|w) & = \frac{e^{\zeta sw}}%
                        {\sum_{s''\in\Omega}e^{\zeta s''w}}
                        \label{eq:binary p(s|w)}
  \end{align}}
\end{minipage}
\begin{minipage}[t]{0.5\textwidth}
  {\begin{align}
    \pi_\mu(a|s)  & = \frac{e^{\mu as}}{\sum_{a'\in\Omega}e^{\mu a's}}
                  \label{eq:binary policy}\\
         p_\tau(w) & = \frac{e^{\tau w}}%
                       {\sum_{w''\in\Omega}e^{\tau w''}}
                       \label{eq:binary model p(w)}
  \end{align}}
\end{minipage}\newline
where all random variables are from the same binary alphabet, i.e.~$w',w,a,s,s'
\in \Omega = \{-1,1\}$. The model is parametrisable by the variables
$\phi,\psi,\zeta,\mu,\tau \in\mathbb{R}^+$, which control how deterministic the
kernels are. If we consider the policy $\pi_\mu(a|s)$ which is controlled by the
parameter $\mu$, then we see from Equation~\ref{eq:binary policy} that $\mu=0$
results in $\pi_\mu(a|s)=\nicefrac12$ for all $a,s\in\Omega$. The policy is
completely random, because both actions $a=1$ and $a=-1$ occur with equal
probability, independent of the current sensor value $s$. On the other hand, if
we set $\mu\gg0$, then the policy changes to a Dirac measure on the sensor and
actuator states, i.e. $\pi_\mu(a|s)=\delta_{as}$, where $\delta_{xy}=1$ if $x=y$ and
zero otherwise. The parameter $\mu$ allows a linear transition between these
two cases.

\begin{figure}[ht]
  \begin{center}
    \includegraphics[width=0.3\textwidth]{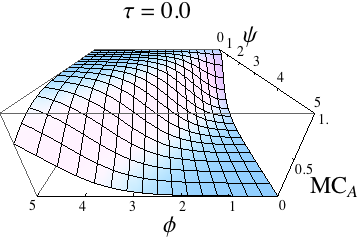}
    \hfill
    \includegraphics[width=0.3\textwidth]{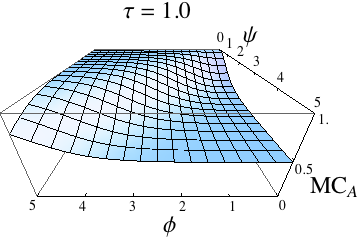}
    \hfill
    \includegraphics[width=0.3\textwidth]{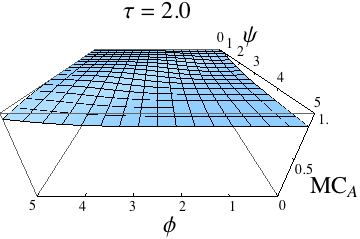}
    \\
    \includegraphics[width=0.3\textwidth]{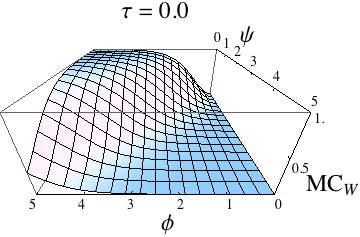}
    \hfill
    \includegraphics[width=0.3\textwidth]{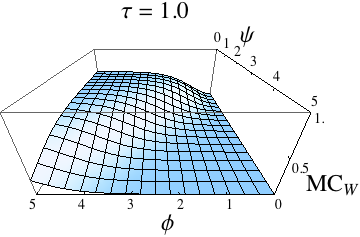}
    \hfill
    \includegraphics[width=0.3\textwidth]{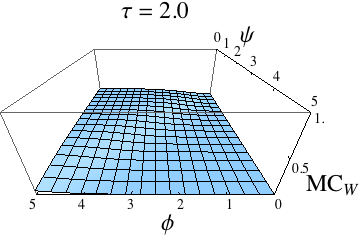}
    \caption{Numerical results for the measures \MCA and \MCW and the binary
      model. The three figures in the first row show the results for \MCA with
      increasingly deterministic policy $\pi$ (from left to right). The second
      row shows the results for \MCW, also with increasingly deterministic
      policy. For all plots, the base axes are given by world kernel parameters
      $\psi$ and $\phi$. The plots confirm the considerations discussed in the
      text below. 
    }
    \label{fig:binary model plots mc mc'}
  \end{center}
\end{figure}

\begin{figure}[h]
  \begin{center}
    \includegraphics[width=0.3\textwidth]{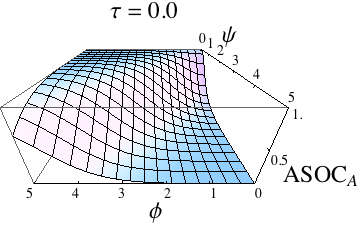}
    \hfill
    \includegraphics[width=0.3\textwidth]{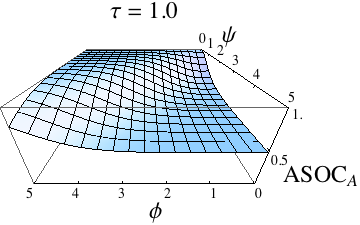}
    \hfill
    \includegraphics[width=0.3\textwidth]{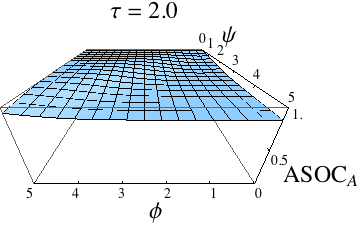}
    \\
    \includegraphics[width=0.3\textwidth]{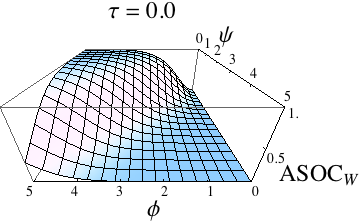}
    \hfill
    \includegraphics[width=0.3\textwidth]{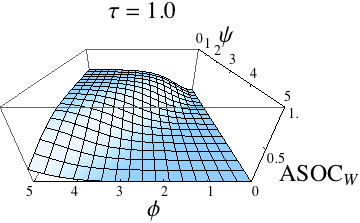}
    \hfill
    \includegraphics[width=0.3\textwidth]{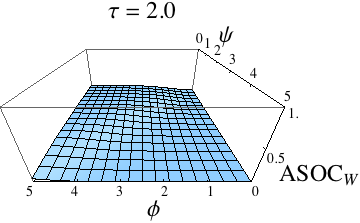}
  \caption{Comparison of the two measures \ASOCA and \ASOCW. The results confirm
    that the measures are intrinsic adaptations of the measures \MCA and \MCW
    (see Figure \protect\ref{fig:binary model plots mc mc'}).}
  \label{fig:binary model plots mc1 mc2}
  \end{center}
\end{figure}

\begin{figure}[h]
  \begin{center}
    \includegraphics[width=0.3\textwidth]{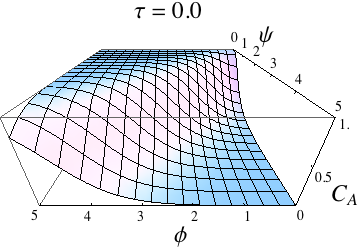}
    \hfill
    \includegraphics[width=0.3\textwidth]{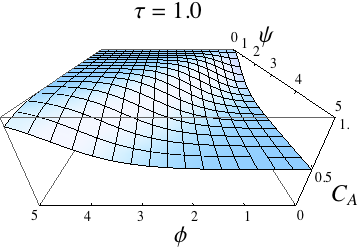}
    \hfill
    \includegraphics[width=0.3\textwidth]{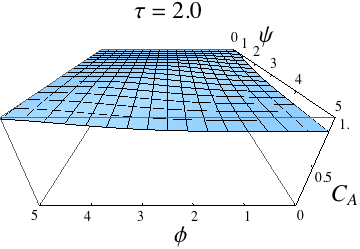}
    \\
    \includegraphics[width=0.3\textwidth]{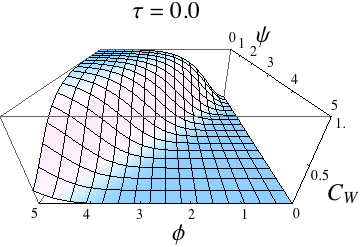}
    \hfill
    \includegraphics[width=0.3\textwidth]{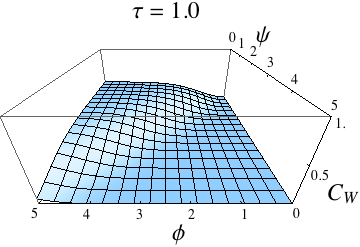}
    \hfill
    \includegraphics[width=0.3\textwidth]{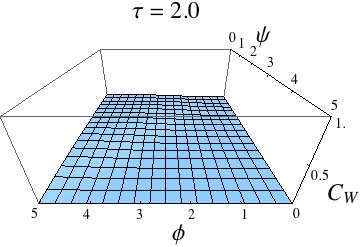}
  \end{center}
  \caption{Comparison of the two measure \CA and \CW. The results confirm
    the measures are intrinsic adoptions of the measures \MCA and \MCW (see
    Figure \protect\ref{fig:binary model plots mc mc'}).}
  \label{fig:binary model plots mc1' mc2'}
\end{figure}

For simplicity, but without loss of generality, the following two assumptions
are made. First, it is assumed that all world states $w\in\Omega$ occur with
equal probability, i.e. $p(w=1) = p(w=-1) = \nicefrac12$. Furthermore, we assume
a deterministic sensor, i.e.~$\zeta\gg1\Rightarrow p(s|w) = \delta_{sw}$, which
means that the sensor is a copy of the world state. The first assumption does
not violate the generality, because it only assures that the world state itself
does not already encode some structure, which is propagated through the
sensori-motor loop. Second, in a reactive system, as it is shown in Figure~\ref{fig:binary model}, the sensor state $S$ and $A$ could be reduced to a
common state, with a new generative kernel $\gamma(a|w) = \pi(a|s) \circ
\beta(s|w)$. Hence, keeping one of the two kernels deterministic and varying the
other in the experiments below, does not effect the generalisation properties of
this model. This leaves three open parameters $\psi,\phi$, and $\mu$, against
which the morphological computation measures are validated (see Figures~\ref{fig:binary model plots mc mc'}, \ref{fig:binary model plots mc1 mc2}, and
\ref{fig:binary model plots mc1' mc2'}).

\begin{figure}[ht]
  \begin{center}
    \includegraphics[height=3cm]{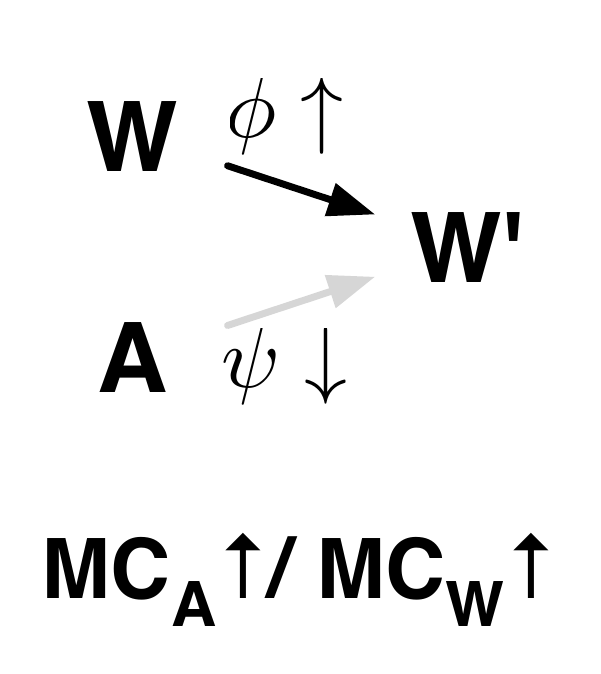}\hspace*{2ex}
    \includegraphics[height=3cm]{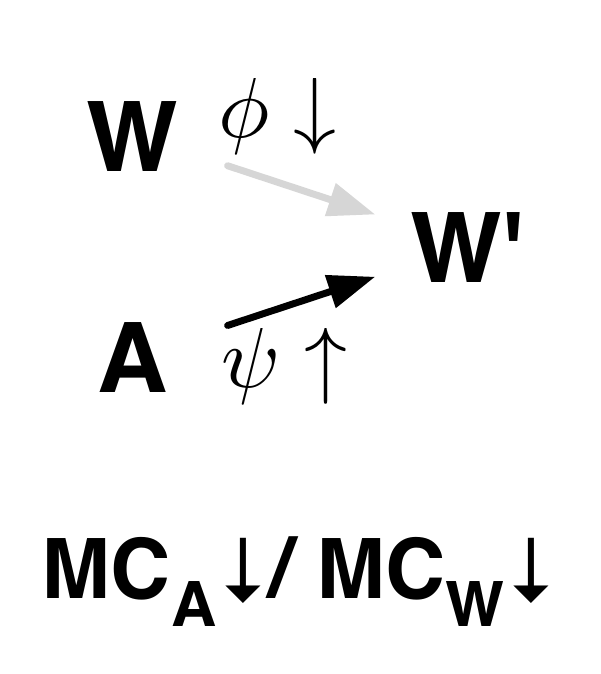}\hspace*{2ex}
    \includegraphics[height=3cm]{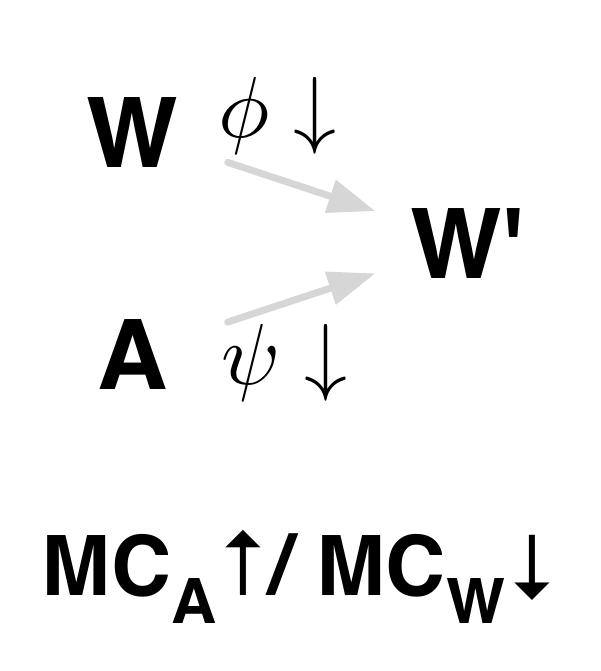}\hspace*{2ex}
    \includegraphics[height=3cm]{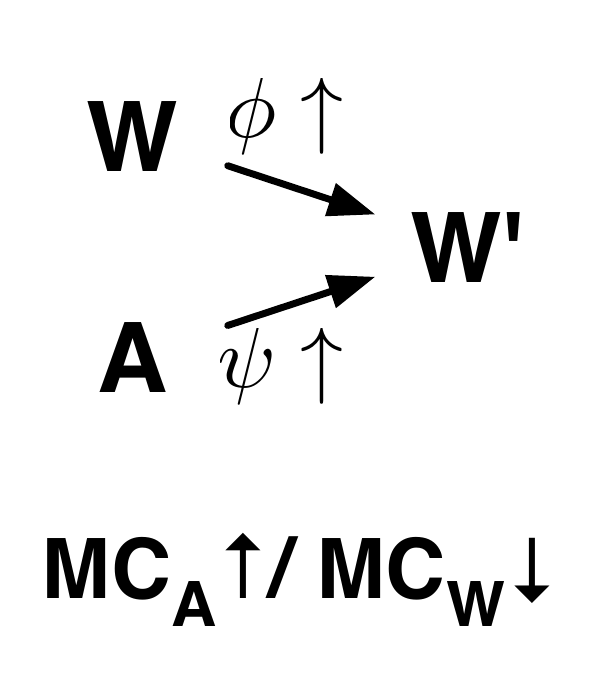}
  \end{center}
  \caption{Visualisation of the four discussed cases of the world model. High
    values are indicated by an arrow pointing upwards ($\uparrow$), and low
    values with an arrow pointing downwards ($\downarrow$),
    i.e.~$\phi\uparrow,\phi\downarrow$ refers to the case, where $\phi\gg0$ and
    $\phi\approx0$, resulting in a high influence of $W\rightarrow W'$ and low
    influence of $A\rightarrow W'$. The figures are discussed in the text below.}
  \label{fig:four cases}
\end{figure}

To be able to understand if the plots validate the measures, it is necessary to
understand how the parameters affect the morphological computation in this model.
To simplify the argumentation, it is first assumed that the policy is
completely random ($\mu=0$), and hence, the action $A$ is independent of the
previous world state $W$ (this assumption is dropped below). What follows is
that the world transition kernel $\alpha(w'|a,w)$ (see Eq.~\ref{eq:binary world
model}) can be divided into four cases (see
Fig.~\ref{fig:four cases}), which are classified by 
the effect of $\phi$ and $\psi$ on the morphological computation.
To understand how the parameters affect the measure, they are first discussed
with respect to the world kernel $\alpha(w'|w,a)$. This is then related to how
the Kullback-Leibler divergences used in the two definitions (see
Def.~\ref{def:MC_A} and Def.~\ref{def:MC_W}) are affected. The analysis is
conducted with respect to the two concepts \MCA and \MCW. It will start with
the two unambiguous cases. The intrinsic measures are discussed later in this
section.

\paragraph{1st Case: $\phi\gg0$, $\psi\approx0$.}
This case refers to a world, which is only influenced by the last world state, as the
world kernel $\alpha(w'|a,w)$ reduces to a Dirac measure on the current and
next world state (see Fig.~\ref{fig:four cases}):
\begin{align}
  \alpha(w'|w,a) & = \delta_{w'w}.
\end{align}
The result is a high measured morphological computation for both measures
\MCA and \MCW, as $p(w'|w,a) = p(w'|w)$, and hence, $-D(p(w'|w,a)||p(w'|w))$
and $D(p(w'|w,a)||p(w'|a))$ are both maximal.

\paragraph{2nd Case: $\phi\approx0$, $\psi\gg0$.}
This case refers to a world, which is only influenced by the last action, as the world
kernel $\alpha(w'|w,a)$ reduces to a Dirac measure on the action and next world
state (see Fig.~\ref{fig:four cases}):
\begin{align}
  \alpha(w'|w,a) & = \delta_{w'a}.
\end{align}
The result is a low measured morphological computation for both measures \MCA
and \MCW, as $p(w'|w,a) = p(w'|a)$, and hence, $-D(p(w'|w,a)||p(w'|w))$ and
$D(p(w'|w,a)||p(w'|a))$ are both minimal.

\paragraph{3rd Case: $\phi\approx0$, $\psi\approx0$.}
This case refers to a world, in which the next world state is independent of the
previous world state and the previous action, which is described by
\begin{align}
  \alpha(w'|w,a)& = \nicefrac12.
\end{align}
The two measures \MCA and \MCW give different results, because the two
quantifications $-D(p(w'|w,a)||p(w'|w))$ and $D(p(w'|w,a)||p(w'|a))$ are both
zero. Consequently, the two measures lead to $\MCA=1$ and $\MCW=0$.

\paragraph{4th Case: $\phi\gg0$, $\psi\gg0$.}
This case is a mixture of all previous cases, as the world behaves according to
\begin{align}
  \alpha(w'|w,a) & = \left\{\begin{array}{ll}
          \delta_{w'w} & \text{if $w=a$}\\
           \nicefrac12 & \text{otherwise}\end{array}\right..
\end{align}

The fourth case is similar to the third case if the world state $W$ is not
equivalent to the action $A$, because in this case the next world state $W'$ is
independent of
both. We saw from the third case, that this leads to $\MCA=1$ and $\MCW=0$. If
the action $A$ and the world state $W$ are equal, the first measure reduces the
amount of morphological computation, whereas the second measure increases the
amount of morphological computation, leading to $\MCA<1$ and $\MCW>0$.

To complete the analysis of the model before the numerical results are
discussed, the assumption of a random policy is dropped. We now assume a fully
deterministic policy ($\mu\gg0$). In this case, the policy reduces to a Dirac
measure on the world and action state, as the sensor state is a copy of the
world ($\beta(s|w)=\delta_{sw}$), and therefore, $\pi_\mu(a|s) = \delta_{as} =
\delta_{as}\delta_{sw}=\delta_{aw}$. Is also follows that $p(w'|a) = p(w'|w)$.
For the two measures, it follows that:
\begin{align}
  \MCA & = 1 - \frac{1}{\ln{|W|}} D(p(w'|w,a)||p(w'|w)) & 
  \MCW & = \frac{1}{\ln{|W|}} D(p(w'|w,a)||p(w'|a)) \\
       & = 1 - \frac{1}{\ln{|W|}} D(p(w'|w)||p(w'|w)) & 
       & = \frac{1}{\ln{|W|}} D(p(w'|w)||p(w'|w)) \\
       & = 1  & 
       & = 0 \label{eq:mca vs mcw}
\end{align}
The effect, that a deterministic policy leads to different results for both
concepts will occur in the following experiments. It will be discussed in detail
in the next section (see Sec.~\ref{sec:discussion}).

The four cases are identical with the four corners of the $\phi$--$\psi$ plane
in Figure~\ref{fig:binary model plots mc mc'}, and the four
cases are well-reflected in the plots. The plots also visualise the difference
of the two concepts for the last two cases ($\phi\approx5$, $\psi\approx5$ and 
$\phi\approx0$, $\psi\approx0$ in Fig.~\ref{fig:binary model plots mc mc'}).

Next, this section now applies the four intrinsic measures to the binary
model. The results are shown for the variations of
the three parameters $\mu,\psi,\phi$ in Figure~\ref{fig:binary model plots mc1
mc2} and Figure~\ref{fig:binary model plots mc1' mc2'}.

The intrinsic measures require the probability distributions $p(s)$ and
$p(s'|s,a)$, which can be calculated from the Equations~\ref{eq:p(w'|w,a)}
to~\ref{eq:binary model p(w)} in the following way:
\begin{align}
    p(s) & = \sum_{w\in\Omega} \beta(s|w) p(w) \label{eq:binary sensor}\\
   p(s'|s,a) & = \sum_{w,w'\in\Omega} p(s',w',w|s,a) = \sum_{w',w\in\Omega}
                  \frac{p(s',w',w,s,a)}{p(s,a)} \\
             & = \sum_{w',w\in\Omega} \frac{p(s',w',w,s,a)}{\pi(a|s)p(s)}\\
             & = \sum_{w',w\in\Omega} \beta(s'|w') \alpha(w'|w,a) \beta(s|w)
                  p(w)\frac{\pi(a|s)}{\pi(a|s)p(s)}\\
             & = \sum_{w',w\in\Omega} \beta(s'|w') \alpha(w'|w,a) \beta(s|w)
                  p(w)\frac{1}{\sum_{w''\in\Omega}\beta(s|w'')p(w'')}
                  \label{eq:binary world model}
\end{align}

The plots reveal that both measures in the first concept and both measures in
the second concept show very similar results compared to the formalisation of
the concepts (compare Fig.~\ref{fig:binary model plots mc mc'} with
Fig.~\ref{fig:binary model plots mc1 mc2} and Fig.~\ref{fig:binary model plots
  mc mc'} with Fig.~\ref{fig:binary model plots mc1' mc2'}). This can be
expected because the sensor distribution is equivalent to the world
distribution due to the Equation~\ref{eq:binary sensor} and because the sensor
kernel was set to be a Dirac measure on the world ($\beta(s|w)=\delta_{sw}$).
Nevertheless, the plots show that the intrinsic adaptations of the formalisation
of the concepts capture what the concepts specify.

%%%%%%%%%%%%%%%%%%%%%%%%%%%%%%%%%%%%%%%%%%%%%%%%%%%%%%%%%%
%%%%%%%%%%%%%%%%%%%%%%%%%%%%%%%%%%%%%%%%%%%%%%%%%%%%%%%%%%
%%%   ROTATOR
%%%%%%%%%%%%%%%%%%%%%%%%%%%%%%%%%%%%%%%%%%%%%%%%%%%%%%%%%%
%%%%%%%%%%%%%%%%%%%%%%%%%%%%%%%%%%%%%%%%%%%%%%%%%%%%%%%%%%

\subsection{Rotator}
\label{sec:rotator}

The previous section verified the measures in a very simple, yet illustrative
binary model of the sensori-motor loop. It was shown that all measures
produce the desired output for variations on the transition probabilities of
the world and the policy. This section applies the intrinsic measures to an experiment
that is inspired by common examples of previously discussed systems that show
high and low morphological computation.

The Dynamic Walker \cite{Wisse2004Essentials-of-dynamic} is designed to emulate
the natural walking of humans. One characteristic is that only half of the leg
movement involved in the walking behaviour is actively controlled by the brain
\cite{Pfeifer2006How-the-Body}. It is the stance phase, i.e.~the time during
which the foot touches the ground and moves the rest of the body forward, which
is fully controlled. The swing phase, i.e.~the time in which the leg swings
forward before another stance is initiated, is only partially controlled.
One may say that the body lets loose and gravity takes over. This is exactly
what the Passive Dynamic Walker highlights, as it only exploits its body and the
environment, i.e.~the slope and gravity to produce a natural walking behaviour.

The world's most advanced humanoid robot
\citep[quoted from][]{American-Honda-Motor-Co.-Inc2013Honda} Asimo is an example of a system,
that does not show any morphological computation as it is discussed in this
work. The trajectory of each part of the morphology is carefully controlled
during the stance and swing phase at all times. This is also the reason, why the
motion of Asimo does not appear natural although it is very smooth.

We are now presenting a simple experiment, that allows us to vary the amount
of exposed morphological computation between the two examples discussed above. For this
purpose we chose a pendulum, which can rotate freely around its anchor point.
The task of our controller is to consistently rotate the pendulum clockwise. The
Asimo case is approximately given, if the angular velocity of the pendulum is
controlled at every instance in time, whereas the Dynamic Walker case is
approximated if the angular velocity of the pendulum is only altered for large
deviations of the current angular velocity from the target velocity. This will
be more clear, after the equations have been presented. The pendulum is
modelled by the following equation:
\begin{align}
  0 & = ml\ddot\theta(t) + \gamma l \dot\theta(t) + m g\sin(\theta(t))-f(t),
  \label{eq:simulation}
\end{align}
where $f(t)$ is the force imposed by the controller (see below), $\gamma$ is the
friction coefficient, $\theta(t)$ is the current angle of the pendulum, $m$ is
its point mass which is located from the center by the length $l$. The Equation~\ref{eq:simulation} was numerally solved for $t=0.0s\ldots0.01s$ using NDSolve
method of Mathematica 8.0 \cite{Research2010Mathematica-Edition:-Version}. The
actions $f(t)$ were modified only for $t=0.01, 0.02, \ldots, T$, and kept constant
while the Equation~\ref{eq:simulation} was solved. This refers to a behaviour
update frequency of 100Hz. The controller is defined by the following set of
equations:
\begin{align}
  s(t) & = \dot\theta(t)+u(-\eta,\eta)\dot\Theta \label{eq:rotator controller 1}\\
    g\left(t\right)      & = \underbrace{\dot\Theta - s(t)}_{\text{1. 
        measured error}} -
  \underbrace{\sgn{s(t)}\,\beta}_{\text{2. reduction due to bias}} +
  \underbrace{\sgn{s(t)}F_\mathrm{min}}_{\text{3. minimial force strength}}
    \label{eq:rotator controller 2}\\
  f(t) & = \left\{
    \begin{array}{ll}
      g\left(t\right)\Big|^{+1}_{-1} \, \cdot F_{\mathrm{max}} & 
      \text{if\ } \left|\dot{\Theta} - s(t)\right| \geq \beta\\
      0 & 
      \text{if\ } \left|\dot{\Theta} - s(t)\right| < \beta\\
    \end{array}\right.. \label{eq:rotator controller 3}
\end{align}
The function of the controller is explained along the three equations
Eq.~\ref{eq:rotator controller 1} to Eq.~\ref{eq:rotator controller 3}. The
first Equation (see Eq.~\ref{eq:rotator controller 1}) adds uniformly
distributed noise $u(-\eta,\eta)\dot\Theta$ to the sensed angular velocity
$\dot\theta(t)$, which is then presented as sensor value $s(t)$ to the
controller. It was discussed in the previous section that a deterministic reactive 
system prevents the possibility to distinguish between the
information flow of $A\rightarrow W'$ and $W\rightarrow W'$ as $A$ becomes a
deterministic function of $W$, and hence, $A$ and $W$ are equivalent with respect
to our measures. Therefore, noise is added to the sensors, such that $S_t$ is
not deterministically dependent on $W_t$. The Equation~\ref{eq:rotator
controller 2} determines
the strength of the response of the controller as a function of the difference
of the sensor value $s(t)$ and the target angular velocity $\dot\Theta$ (see
first term in Eq.~\ref{eq:rotator controller 2}). This function is only executed
if the difference of the sensor value to the goal is larger than a threshold
value $\beta$. Hence, to ensure sensitivity, the threshold value is subtracted
from the absolute value of the response (see second term in Eq.~\ref{eq:rotator
controller 2}). The third term in the Equation~\ref{eq:rotator controller 2}
ensures a minimal response strength, and hence, a minimal effect of the action
$A_t$ on the next sensor value $S_{t+1}$.

All parameters shown in the Equations~\ref{eq:simulation}--\ref{eq:rotator
controller 3} were evaluated systematically. The most distinctive
results were found for the following values, which is why they were chosen for
presentation here:
\begin{align}
  F_\mathrm{max}  & = 10.0 &
  \beta           & \in [0.0,2.0] & 
  \dot\Theta      & = 2\pi & 
  m               & = 1.0 \\
  F_\mathrm{min}  & = 0.25 &
  \eta            & \in [0.0,0.5] &
  \gamma          & = 0.0 & 
  g               & = 9.81 & 
  T                = 5000
\end{align}

\begin{figure}[t]
  \begin{center}
    \bgroup
      \def\arraystretch{0}
      \setlength{\tabcolsep}{1mm}
      \begin{tabular}{cp{1mm}cp{1mm}cp{1mm}cp{1mm}cccc}
        \multicolumn{2}{c}{\includegraphics[width=0.225\textwidth]{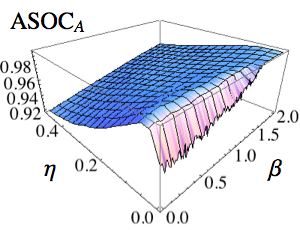}} &
        \multicolumn{2}{c}{\includegraphics[width=0.225\textwidth]{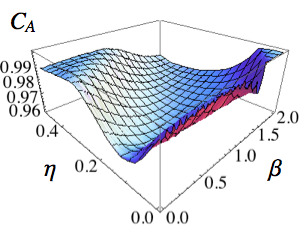}} \hspace*{2ex}&
        \multicolumn{2}{c}{\includegraphics[width=0.225\textwidth]{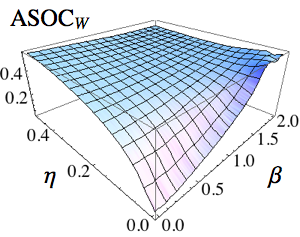}} &
        \multicolumn{2}{c}{\includegraphics[width=0.225\textwidth]{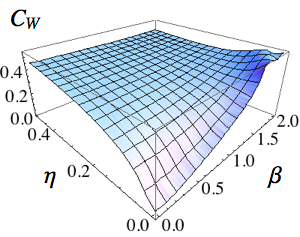}} \\
        \includegraphics[width=0.15\textwidth]{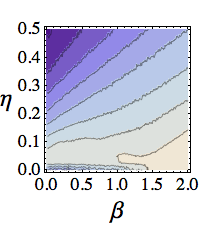} &
        \includegraphics[height=3cm]{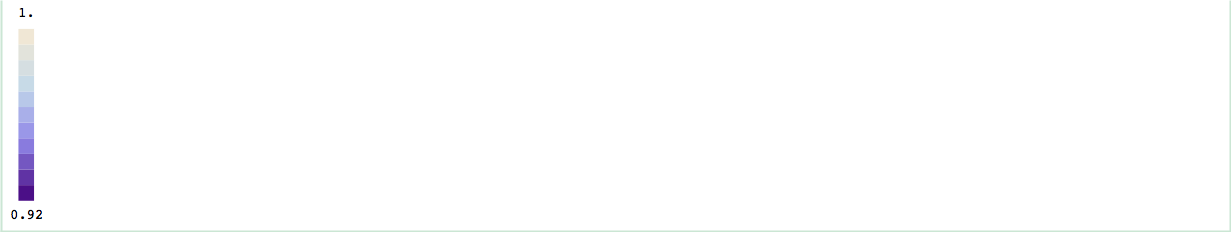} &
        \includegraphics[width=0.15\textwidth]{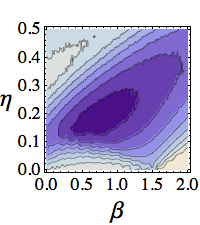} &
        \includegraphics[height=3cm]{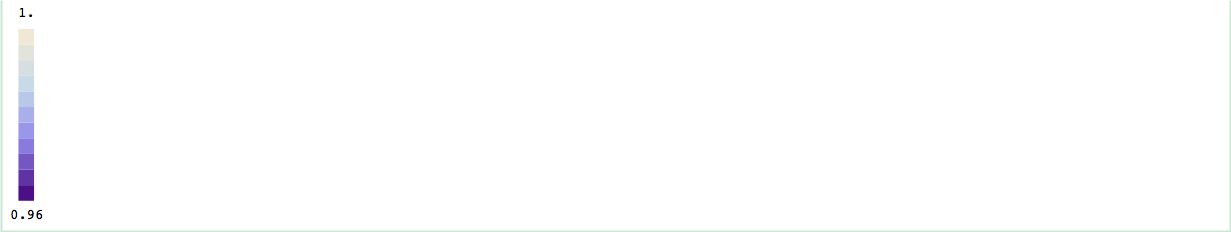} &
        \includegraphics[width=0.15\textwidth]{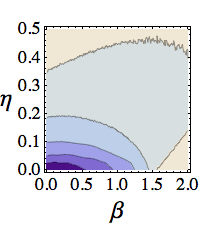} &
        \includegraphics[height=3cm]{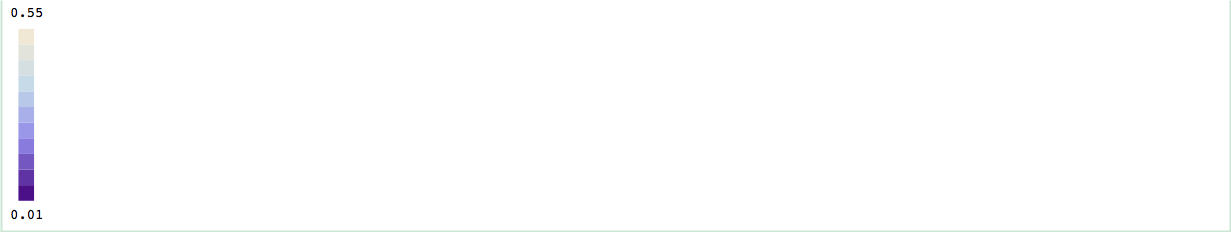} &
        \includegraphics[width=0.15\textwidth]{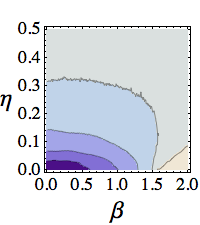} &
        \includegraphics[height=3cm]{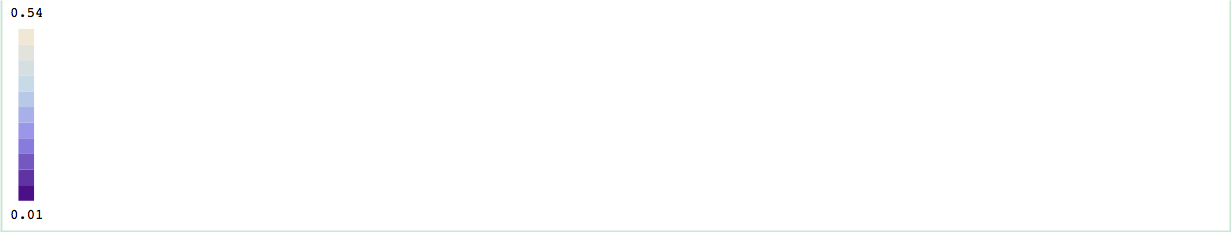}
      \end{tabular}
    \egroup
  \end{center}
  \caption{Results of the intrinsic measure in the rotating pendulum experiment.
  From left to right \ASOCA, \CA, \ASOCW, and \CW. The plane in each plot is
  defined by the noise $\eta$ and the threshold parameter $\beta$. The values
  are averaged over 10 runs, for $\eta\in\{0, 0.025. 0.05,\ldots,0.5\}$ and
  $\beta\in\{0,0.01,0.02,\ldots,2.0\}$. The
  results are discussed in the text below.}
  \label{fig:rotator plots}
\end{figure}

\begin{figure}[h]
  \begin{center}
    \begin{tabular}{cc}
      \includegraphics[width=0.475\textwidth]{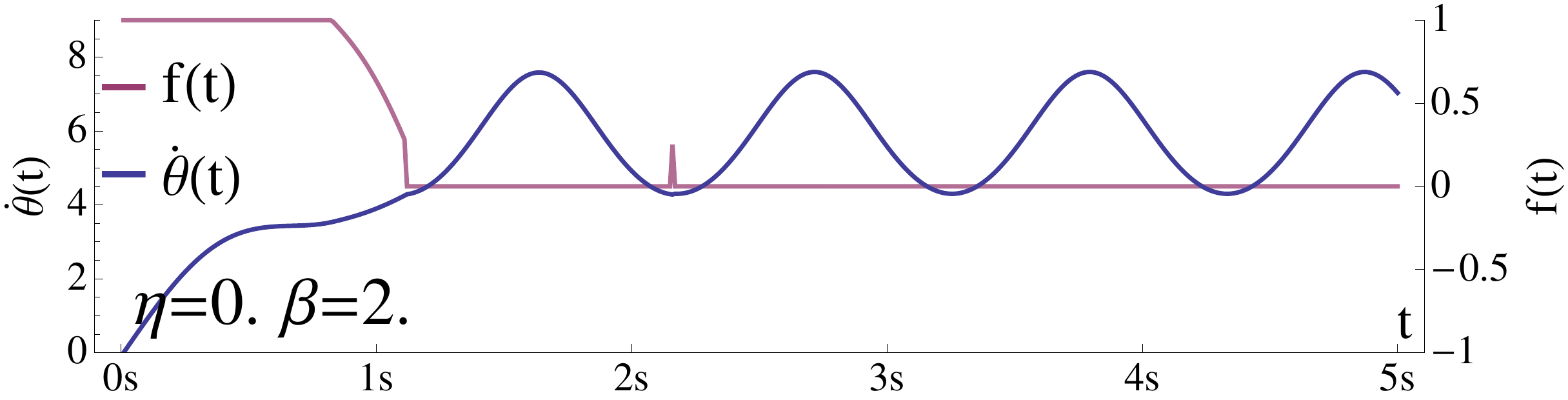} &
      \includegraphics[width=0.475\textwidth]{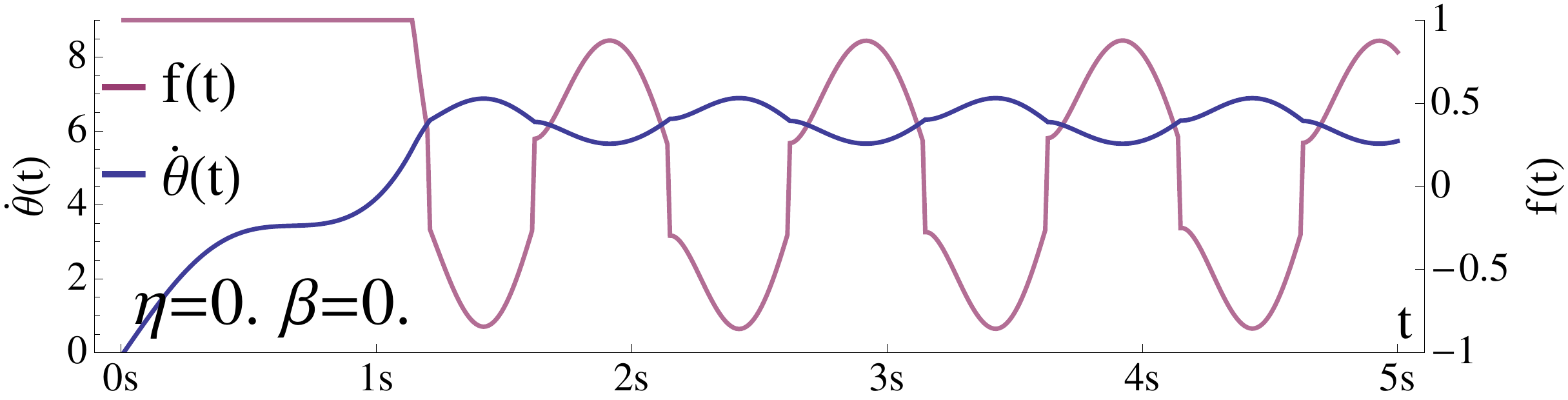}\\
      {\footnotesize \ASOCA: 0.99} \
      {\footnotesize \CA: 1.00}   \
      {\footnotesize \ASOCW: 0.54}  \
      {\footnotesize \CW: 0.54}    &
      {\footnotesize \ASOCA: 0.96} \
      {\footnotesize \CA: 0.99}    \
      {\footnotesize \ASOCW: 0.01} \
      {\footnotesize \CW: 0.01} 
    \end{tabular}\bigskip\\

    \begin{tabular}{cc}
      \includegraphics[width=0.475\textwidth]{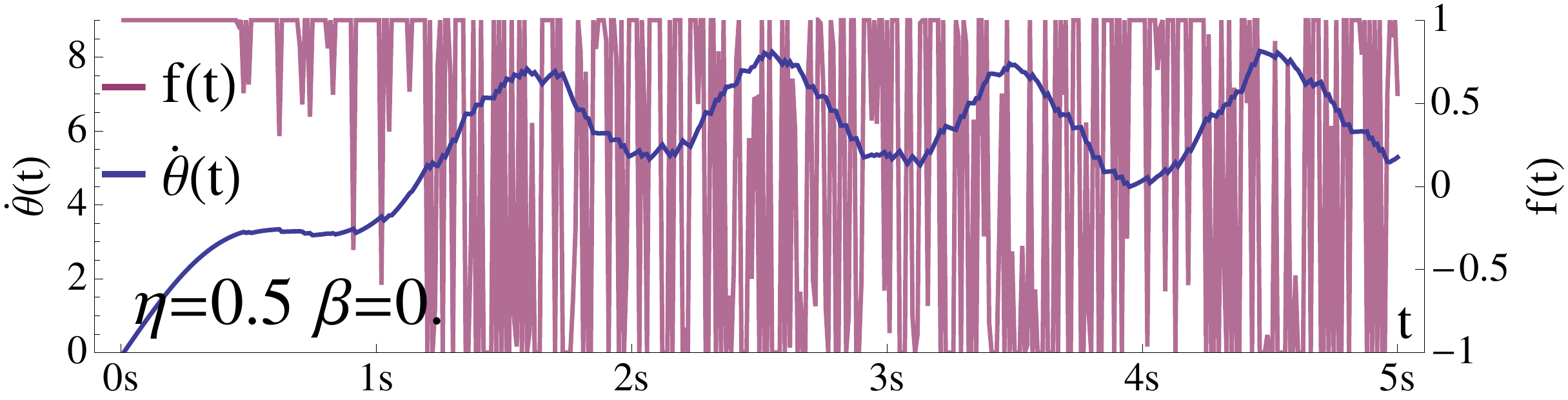} &
      \includegraphics[width=0.475\textwidth]{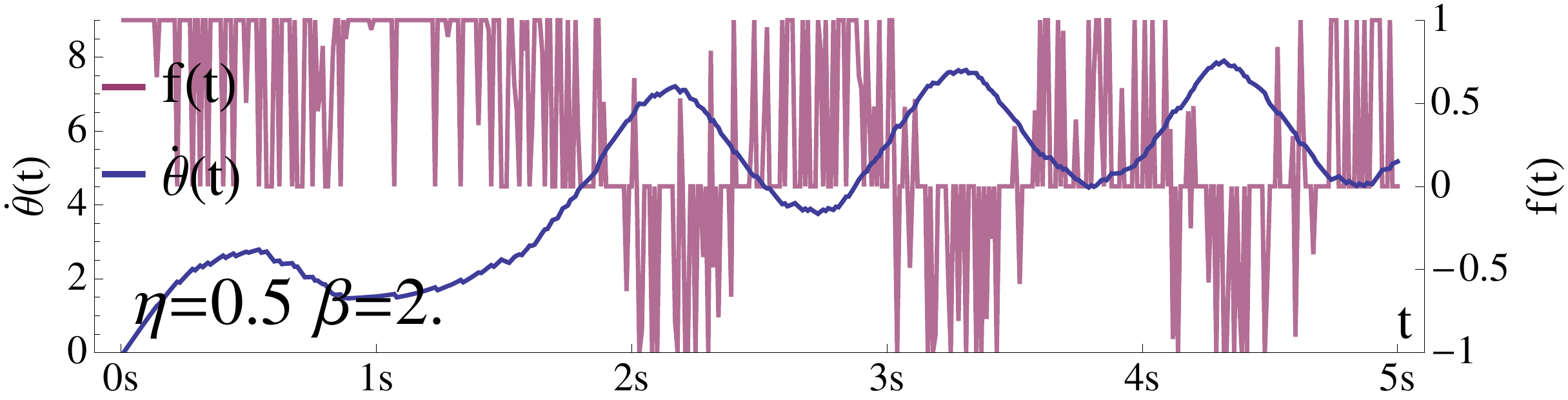} \\
      {\footnotesize \ASOCA: 0.92} \
      {\footnotesize \CA:    0.99} \
      {\footnotesize \ASOCW: 0.46} \
      {\footnotesize \CW:    0.55} &
      {\footnotesize \ASOCA: 0.96} \
      {\footnotesize \CA:    0.97} \
      {\footnotesize \ASOCW: 0.45} \
      {\footnotesize \CW:    0.51}
    \end{tabular}\bigskip\\
  \end{center}
  \caption{Transients. Please see the legend to each plot and read the text
    below for a discussion. The values were averaged over 100 runs.}
  \label{fig:rotator transients}
\end{figure}

The results are presented in two figures (see Fig.~\ref{fig:rotator plots} and
Fig.~\ref{fig:rotator transients}). The Figure~\ref{fig:rotator plots} shows
three-dimensional plots in which the plane is defined by the noise factor $\eta$ and
the threshold $\beta$, and for which the height $z$ is given by the measurements
of the denoted measure. The Figure~\ref{fig:rotator transients} shows the
transients of the sensor values $s(t)$ (see Eq.~\ref{eq:rotator controller 1})
and the normalised action $g(t)\big|_{-1}^{+1}$ (see Eq.~\ref{eq:rotator
controller 2} and Eq.~\ref{eq:rotator controller 3}) for the four extremal
points of the plots in Figure~\ref{fig:rotator plots}. Therefore,
the five plots in Figure~\ref{fig:rotator transients} show the transients for
the configurations $(\eta,\beta)\in\{(0,0), (0.0,2.0), (0.5,0.0), (0.5,2.0)\}$.
The discussion of the results in the following paragraphs refers
to these two figures and follows the ordering of the transient plots in Figure~\ref{fig:rotator transients}. All results were obtained by sampling the world
model $p(s'|s,a)$, policy $p(a|s)$ and sensor distribution $p(s)$ from the
data stream of 5000 behaviour updates (see Eq.~\ref{eq:simulation}). The
sampling was performed according to
\cite{Zahedi2010Higher-coordination-with} (it is additionally also briefly
described in Section~\ref{app:sampling} of the appendix). The sensor values were
binned in the interval $S_t\in[0,8]$ and the actions were binned in the interval
$A_t\in[-1,1]$ with 30 bins each.

The first discussed configuration is $(\eta,\beta)=(0,2.0)$. The transients show
a clear picture, as no action of the controller is performed other than $f(t)=0$
after about 2 seconds. Consequently, for this configuration
any measure should result in maximal morphological computation leading to a
value close or equal to one, as the inertia, and hence, the world $W$ is the
only cause for the observed behaviour. We see that both measures of the first
concept (\ASOCA and \CA) deliver a value close or equal to 1 (all values in the Figure~\ref{fig:rotator transients} are rounded to the second decimal place and are
averaged over 100 runs). The two
measures of the second concept (\ASOCW and \CW) show their maximal values 
for this configuration at approximately 0.54 (maximal refers to all
configurations shown in Figure~\ref{fig:rotator plots}). This
again highlights the differences of the two concepts. The first concept reduces
the amount of morphological computation by the measured effect of the action $A$
on the next world state $W'$, measured through $S'$, whereas the second concept
increases the morphological computation by the measured effect of $W$ on $W'$,
captured by $S$ and $S'$. This explains, why both concepts show different
maximal values for this configuration.

The second discussed configuration is given by $(\eta,\beta)=(0,0)$. We 
expect no morphological computation, because a threshold of
$\beta=0$ means that the pendulum is controlled at every time step $t$, and no
noise on the sensor values $\eta=0$ means that the action is only dependent on
the actual angular velocity $\dot\theta(t)$. We
see that in this case, both concepts lead to very different results. The second
concept matches our intuition better, as both  measures (\ASOCW and \CW) deliver
values close to zero. The measures in the first concept result in values close to
one, as the action $A$ is deterministically dependent on the world state $W$,
captured by $S$, and hence, the effect of $A\rightarrow W'$ is fully determined by
the effect of $W\rightarrow A$. Consequently, in the first concept, the current world
state $W$ almost fully determines the next world state $W'$, which leads to the
high values. We want to point out that there is no preference
to any one of the two results. Both seem equally valid at the current
stage of the discussion. This is strong evidence that both concepts capture
important aspects, but that a clear picture cannot be presented yet. We will 
discuss this in detail in the next section.

The other two configurations show that a noisy controller output is filtered
by the inertia of the pendulum, resulting in an angular velocity approximating
the target velocity. The amount of noise $\eta$ increases the measured
morphological computation as well as the higher threshold value $\beta$. Both
results are consistent with our expectations.

%%%%%%%%%%%%%%%%%%%%%%%%%%%%%%%%%%%%%%%%%%%%%%%%%%%%%%%%%%%%%%%%%%%%%%%%%%%%%%%%
%%%%%%%%%%%%%%%%%%%%%%%%%%%%%%%%%%%%%%%%%%%%%%%%%%%%%%%%%%%%%%%%%%%%%%%%%%%%%%%%
%%%   DISCUSSION
%%%%%%%%%%%%%%%%%%%%%%%%%%%%%%%%%%%%%%%%%%%%%%%%%%%%%%%%%%%%%%%%%%%%%%%%%%%%%%%%
%%%%%%%%%%%%%%%%%%%%%%%%%%%%%%%%%%%%%%%%%%%%%%%%%%%%%%%%%%%%%%%%%%%%%%%%%%%%%%%%

\section{Discussion}
\label{sec:discussion}

\citet{Pfeifer1999Understanding-intelligence} state that
\emph{one problem with the concept of morphological computation is that while
intuitively plausible, it has defied serious quantification efforts}.
From the experiments presented in the previous section we can now understand where
this problem rises from. In the context of a deterministic world, a 
deterministic embodiment, and a deterministic reactive policy one cannot
distinguish between the effect of the world $W$ and the effect of the action $A$
on the next world state $W'$ as the action $A$ is given
by a deterministic function of $W$. Hence, the two states can be subsumed to a
single state, which results in the obtained difference of the measures in the
two concepts (see Eq.~\ref{eq:mca vs mcw}). We will study this with respect to a
\emph{Gedankenexperiment} by \citet{Braitenberg1984Vehicles}. In his book,
Braitenberg discusses different vehicles, which show different behaviours due to
a very simple sensor-motor coupling. The vehicle 3 is of such kind. It has two
\begin{figure}[h]
  \begin{center}
    \includegraphics[width=0.75\textwidth]{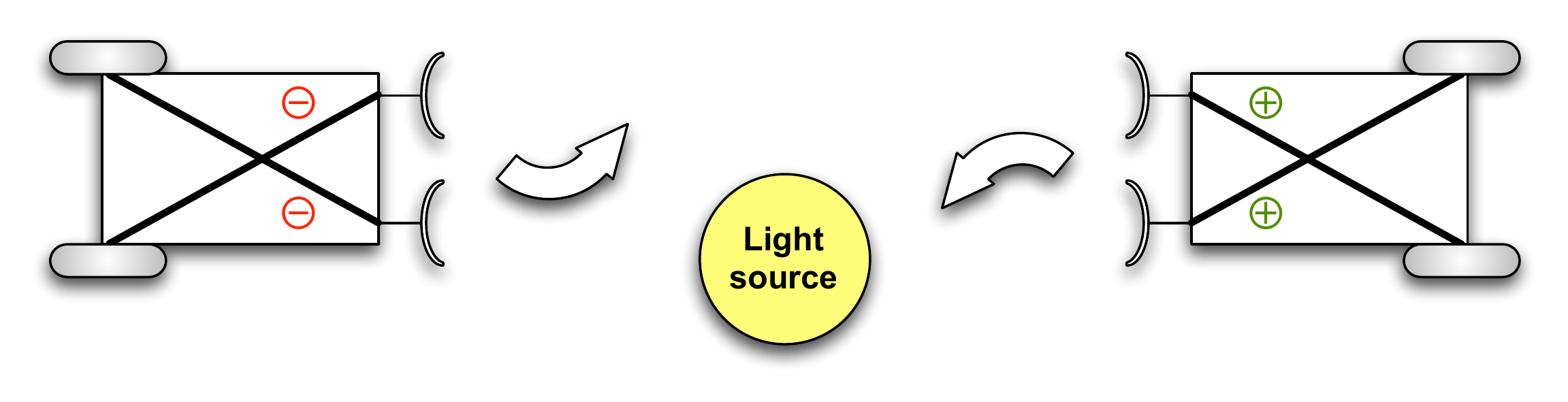}
  \end{center}
  \caption{Braitenberg Vehicle 3. The Figure shows two Braitenberg vehicles,
    each of them is equipped with two light sensors and two actuators. The
    polarity of the sensor-motor couplings determines the behaviour. The
    Braitenberg vehicle on the left-hand side is repelled by light, whereas the
    Braitenberg vehicle on the right-hand side is attracted by light.}
  \label{fig:BV 3}
\end{figure}
sensors that can detect light sources and two actuators with which the system
can move and steer. The sensors and motors are cross-coupled (see
Fig.~\ref{fig:BV 3}). Depending on the polarity of the couplings, the vehicle is
either attracted or repelled from a light source. For this experiment, the two
measures result in $\MCA=1$ and $\MCW=0$. Which one is considered to be correct
depends on how the policy is separated from the body. One may argue, that
the coupling and the coupling strength defines the policy. In this case, one has
to conclude that there is low morphological computation, as the next world state
is mostly determined by the policy. This favors the measure \MCW. One may also
argue that the couplings between the sensors and actuators are of such
simplicity, that they can be considered as part of the body, as the sensor
signals directly feed to the actuators. In this case, the state of the system is
fully determined by the current sensor state, and hence, the world state. This
is in favor of the first measure \MCA.

It seems that both concepts capture important aspects of morphological
computation, but that a final answer cannot yet be given here, as the discussion
about the Braitenberg vehicle 3 shows. 

{%\color{BlueGreen}
Another important point must be discussed here. All presented quantifications
measure the influence of the world state $W$ on the next world state $W'$ in one
way or the other (best seen in Fig.~\ref{fig:m2 prime concept}). One may argue,
that measuring morphological computation requires to take the control system
into account, as an open-loop control system should show more morphological
computation compared to a closed-loop control system, because the morphology
becomes more relevant. How closed-loop and
open-loop systems relate, was first investigated by
\citet{Touchette2004Information-theoretic-approach-to-the-study}, who measured
how much information a control system acquires from the environment. That this
complementary concept does not well-capture morphological computation is shown
by the following example. A manufacturing robot working in an industrial
assembly line is not associated with high morphological computation, because it
is especially constructed and programmed to compensate for any uncontrolled
effects resulting from the morphology. The
trajectory of the tool is preprogrammed to the extend that any human entering
its working space during operation is risking serious injuries. This is because
this type of robots are not equipped with sophisticated sensors and mainly
controlled in an open-loop paradigm. Morphological computation is the
information that the \emph{environment} processes as opposed to the information
that the \emph{control system} acquires
\citep{Touchette2004Information-theoretic-approach-to-the-study} or processes
from the environment.

We previously discussed the connection of morphological computation to the
Information Bottleneck Method. Here we discussed the connection open- and
closed-loop control to our measures. Both are dealing with the mutual
information of the current world state $W$ and the current action $A$ in
different ways. Investigating the connection between the Information Bottleneck
Method, Touchette and Lloyd's approach and morphological computation seems to be
a very promising approach to generate new insights about the sensori-motor loop.
Unfortunately, it is beyond the scope of this work.}

\section{Conclusions}
\label{sec:conclusions}

This work began with an introduction of morphological computation in the context
of embodied artificial intelligence. It was concluded, that the complexity of
the controller or brain should not be the primary ingredient for a measure of
morphological computation. Instead, the influence of the action of an agent and
the influence of the last world state need to be compared. Two concepts to
measure morphological computation were then discussed and formalised, from which
several intrinsic adaptations were derived. The different measures were
evaluated in two experiments, which showed the conceptual difficultly in
measuring morphological computation. For any fully deterministic reactive
system, the influence of the action and the influence of the world are not
easily separated, which leads to different results for the two concepts. As this
only occurs in this very special case, {%\color{BlueGreen}
  i.e.~when no noise is present in the
entire loop,} we propose to use the different results as an indication that
the observed behaviour may be due to such a deterministic reactive system. 

From the derivation of the concepts it is clear that
the world states $W$ and $W'$ may not only include an individual's morphology and
\emph{Umwelt} but also those of other agents. Hence, our measures also apply to
quantify collective behaviours, as e.g.~stigmergy, flocking, etc. Discussing
this is beyond the scope of this work but subject to future research.

Two concluding remarks will close this work. First, the presented
concepts each point to important aspects of morphological computation, but are
not the final answer. A final measure will have to combine both concepts into
one. How this can be done is open to future work. Second, although a definite
answer is not yet given, the intrinsic adaptations are good candidates for
self-organisation principles. An embodied system should maximise its
morphological computation in order to minimise its computational requirements.
This may have implications for practical applications in the field of
robotics, as the amount of on-board computation, and therefore, load and
energy requirements could be reduced. In first and ongoing experiments we apply
the intrinsic measures to the learning of locomotion for a six-legged walking
machine in the context of reinforcement learning. The travelled distance is
multiplied with the normalised intrinsic measures to give the overall reward.
Here, practical advantages of the measures in the second concept seem to be
that they start with zero, which means that there is a higher pressure to
exploit the morphology, and that they are more sensitive, which was
already seen in the rotator experiment presented above.

Concluding we state, that
although not finally answered, the proposed measures already give powerful tools
to measure and exploit morphological computation in the context of embodied
intelligence.

\section*{Acknowledgements}
This works was partly funded by the DFG Priority Program 1527, Autonomous Learning.

\appendix
\section{Appendix}

\subsection{Derivation of the Causal Measure 1}
\label{app:deviation of causal measure 1}

The causal information flows used in the Section~\ref{sec:measuring
  morphological computation} (see Fig.~\ref{fig:causal idea
s,s'} and Fig.~\ref{fig:causal idea a,s'}), are
derived by the means of a mutual information measure. The mutual information
of two random variables $X$ and $Y$ is calculated in the following way
\citep{Cover2006Elements-of-Information}:
\begin{align}
  I(X;Y) & = \sum_{x\in\mathcal{X}}\sum_{y\in\mathcal{Y}} p(x,y) \ln\left[
\frac{p(x,y)}{p(x)p(y)}\right]\\
  & = D(p(x,y)||p(x)p(y)) \label{eq:D MI}
\end{align}
This is the Kullback-Leibler divergence of the joint distribution
of $X$ and $Y$ with respect to the product of the marginal distributions of $X$
and $Y$ (see
Eq.~\ref{eq:D MI}). In the case of the causal information flow of $S$ on $S'$,
the measure compares the joint distribution of $S'$ and the intervened $S$,
given by $p(s',\mathrm{do}(s)):= p(s'|\mathrm{do}(s))p(s)$, with the marginal
distribution of $\hat{p}(s')p(s)$, where $\hat{p}(s')$ is the
post-interventional distribution of $S'$ (compare Eq.~\ref{eq:hat p(s) 1} and
\ref{eq:hat p(s) 2}, below). The resulting measure is then given by:
\begin{align}
  CIF(S\rightarrow S') & = \sum_{s'\in\mathcal{S}, s\in\mathcal{S}} p(s',
  \mathrm{do}(s)) \ln
  \left[\frac{p(s',\mathrm{do}(s))}{\hat{p}(s')p(s)}\right]\\
  & = \sum_{s\in\mathcal{S}}p(s)\sum_{s'\in\mathcal{S}} 
  p(s'|\mathrm{do}(s)) \ln
  \left[\frac{p(s'|\mathrm{do}(s))p(s)}{\hat{p}(s')p(s)}\right]\label{eq:hat
  p(s) 1}\\
  & = \sum_{s\in\mathcal{S}}p(s)\sum_{s'\in\mathcal{S}} 
  p(s'|\mathrm{do}(s)) \ln
  \left[\frac{p(s'|\mathrm{do}(s))}
  {\sum_{s''\in\mathcal{S}}p(s'|\mathrm{do}(s''))p(s'')}
\right]
\label{eq:hat p(s) 2}
\end{align}
The causal information flow of $A$ on $S'$, denoted by $CIF(A\rightarrow S')$ is
derived analogously with $p(s',\mathrm{do}(a)):=p(s'|\mathrm{do}(a))p(a)$:
\begin{align}
  CIF(A\rightarrow S') & = \sum_{s'\in\mathcal{S}, a\in\mathcal{A}} p(s',
  \mathrm{do}(a)) \ln
  \left[\frac{p(s',\mathrm{do}(a))}{\hat{p}(s')p(a)}\right]\\
  & = \sum_{a\in\mathcal{A}}p(a)\sum_{s'\in\mathcal{S}} 
  p(s'|\mathrm{do}(a)) \ln
  \left[\frac{p(s'|\mathrm{do}(a))}
  {\sum_{a'\in\mathcal{A}}p(s'|\mathrm{do}(a'))p(a')} 
  \right]\\
  & = \sum_{a\in\mathcal{A}}p(a)CIF(a\rightarrow S')
\end{align}
The Equation~\ref{eq:equivalence} implies 
\begin{align}
  CIF(S\rightarrow S') & = \sum_{s'\in\mathcal{S}}p(s'|\mathrm{do}(s)) \ln
  \left[\frac{p(s'|\mathrm{do}(s))}{\sum_{s'\in\mathcal{S}}p(s'')p(s'|\mathrm{do}(s'))}\right]\\
  & = \sum_{s'\in\mathcal{S}}\left(\sum_{a\in\mathcal{A}}p(a|s)p(s'|\mathrm{do}(s))\right)\ln
  \left[\frac{\sum_{a\in\mathcal{A}}p(a|s)p(s'|\mathrm{do}(a))}{\sum_{s''\in\mathcal{S}}p(s'')
  \sum_{a\in\mathcal{A}}p(a|s)p(s'|\mathrm{do}(a))}\right]\\
  & = \sum_{s'\in\mathcal{S}}\left(\sum_{a\in\mathcal{A}}p(a|s)p(s'|\mathrm{do}(a))\right)\ln
  \left[\frac{\sum_{a\in\mathcal{A}}p(a|s)p(s'|\mathrm{do}(a))}{\sum_{a\in\mathcal{A}}p(a)
  p(s'|\mathrm{do}(a))}\right]\\
  & \leq \sum_{a\in\mathcal{A}}p(a|s)\sum_{s'\in\mathcal{S}}p(s'|\mathrm{do}(a))\ln
  \left[\frac{p(s'|\mathrm{do}(a))}{\sum_{a\in\mathcal{A}}p(a)p(s'|\mathrm{do}(a))}\right]\\
  \Rightarrow    CIF(S\rightarrow S') & \leq
  \sum_{a\in\mathcal{A}}p(a|s)CIF(a\rightarrow S'). \label{eq:inequality}
\end{align}
Summation with respect to $s$ finally yields
\begin{align}
  CIF(S\rightarrow S') & = \sum_{s\in\mathcal{S}} p(s) CIF(s\rightarrow S')\\
                     & \leq \sum_{s\in\mathcal{S}}p(s)\sum_{a\in\mathcal{A}}p(a|s)CIF(a\rightarrow S')\\
                     & = \sum_{a\in\mathcal{A}}p(a)CIF(a\rightarrow S')\\
   \Rightarrow CIF(S\rightarrow S') & \leq CIF(A\rightarrow S')\label{eq:CS smaller CA}
\end{align}
It follows, that the measure for morphological computation can be represented
as KL-divergence in the following form:
\begin{align}
CIF(S\rightarrow S') - CIF(A\rightarrow S') \label{eq:mc2}
     & = - \sum_{s\in\mathcal{S},a\in\mathcal{A}}p(s,a)\sum_{s'\in\mathcal{S}}p(s'|\mathrm{do}(a))\ln\frac{p(s'|\mathrm{do}(a))}{p(s'|\mathrm{do}(s))}\\
     & = -D(p(s'|\mathrm{do}(a))||p(s'|\mathrm{do}(s))) 
\end{align}
The Kullback-Leibler divergence is non-negative
\cite{Cover2006Elements-of-Information}, i.e.~$D(p||q)\geq 0$, which
means that the measure given above (see Eq.~\ref{eq:mc2}) is always negative.

The derivation of the morphological computation measure for non-reactive control
(see Fig.~\ref{fig:sml full causal}) is analogous to the reactive control
\citep{Ay2013An-Information-Theoretic}. Every occurrence of
$s,S,\mathcal{S}$ is replaced by $c,C,\mathcal{C}$, which leads to 
\begin{align}
  CIF(C\rightarrow S') & = \sum_{c\in\mathcal{C}}p(c)\sum_{s'\in\mathcal{S}} 
  p(s'|\mathrm{do}(c)) \ln
  \left[\frac{p(s'|\mathrm{do}(c))}
  {\sum_{c'\in\mathcal{C}}p(s'|\mathrm{do}(c'))p(c')}\right]
\end{align}

\subsection{Sampling world model, policy and input distribution in pendulum
  experiments}
\label{app:sampling}

The probability distributions that were required in the rotator experiment (see
Sec.~\ref{sec:rotator}) were obtained from the data series by the same sampling
method
that we have used in a previous publication
\cite{Zahedi2010Higher-coordination-with}. It is briefly discussed for the world
model below. Its application to the policy and input distribution is straight
forward \cite{Ay2013An-Information-Theoretic}. The sampling starts with 
a uniform distribution and the update is given by:
\begin{align}
  p^{(0)}(s'|s,{a})                  & :=  \frac{1}{|\mathcal{S}|}\nonumber\\
  p^{\left(n_{a}^s\right)}(s'|s,{a}) & :=  \left\{\begin{array}{ll}
  \displaystyle\frac{n_{a}^s}{n_{a}^s+1}p^{(n_{a}^s-1)}(s'|s,{a})+\frac{1}{n_{a}^s+1} 
                  & {{\text{if } {S_{n_{a}^s+1} = s',\, S_n=s,\,A_{n_{a}^s+1}={a}}}}\\[3ex]
  \displaystyle\frac{n_{a}^s}{n_{a}^s+1}p^{(n_{a}^s-1)}(s'|s,a) 
                  & {\text{if } S_{n_{a}^s+1} \not= s',\, S_n=s,\,A_{n_{a}^s+1}={a}}\\[3ex]
  p^{(n_{a}^s-1)}(s'|s,{a})
  & {\text{if } S_{n_{a}^s}\not=s \text{ or } \,A_{n_{_{s,{a}}}+1}\not={a}}
  \end{array}\right. \label{eq:mil world model}
\end{align}

\bibliographystyle{mdpi}
\makeatletter
\renewcommand\@biblabel[1]{#1. }
\makeatother
%\bibliography{morphcomp}

\end{document}